\documentclass[10pt,twocolumn,letterpaper]{article}

\usepackage[pagenumbers]{cvpr} %

\usepackage{soul}
\setuldepth{foobar}

\definecolor{cvprblue}{rgb}{0.21,0.49,0.74}
\usepackage[pagebackref,breaklinks,colorlinks,allcolors=cvprblue]{hyperref}

\usepackage{comment}
\usepackage{multirow}
\usepackage{hyperref}
\usepackage{cleveref}
\usepackage{placeins}
\usepackage{tikz}
\usetikzlibrary{spy,backgrounds}
\usepackage[permil]{overpic}

\usepackage{array}
\usepackage{multirow}
\newcolumntype{M}[1]{>{\centering\arraybackslash}m{#1}}
\newlength\tmpcolwidth

\newcommand{\opacity}{\boldsymbol{o}}
\newcommand{\scale}{\boldsymbol{\sigma}}
\newcommand{\position}{\boldsymbol{\mu}}
\newcommand{\rotation}{\mathbf{R}}
\newcommand{\ra}{\mathbf{v_1}}
\newcommand{\rb}{\mathbf{v_2}}

\newcommand{\rc}{\mathbf{v_3}}
\newcommand{\radiance}{\mathbf{c}}
\newcommand{\ray}{\mathbf{r}}
\newcommand{\origin}{\mathbf{o}}
\newcommand{\rayparam}{t}
\newcommand{\dir}{\mathbf{d}}

\newcommand{\transmittance}{T}

\newcommand{\radiancetexture}{\radiance^{\oplus}}
\newcommand{\radianceplain}{\radiance^{\odot}}
\newcommand{\weight}{\mathbf{w}}

\newcommand{\Weights}{\mathbf{W}}
\newcommand{\Textures}{\mathbf{T}}
\newcommand{\Notextures}{\mathbf{N}}
\newcommand{\Ids}{\mathbf{I}}
\newcommand{\Depths}{\mathbf{D}}

\DeclareMathOperator{\erf}{erf}
\DeclareMathOperator*{\argmax}{arg\,max}

\usepackage{bbding}
\usepackage{pifont}
\def\faChecked{\ding{51}}
\def\faCrossed{\ding{55}}

\definecolor{lightblue}{HTML}{0080FE}
\definecolor{darkblue}{HTML}{000080}

\def\paperID{} %
\def\confName{CVPR}
\def\confYear{2026}

\title{Nexels: \ul{Ne}urally-Te\ul{x}tured Surf\ul{els}
\\ for Real-Time Novel View Synthesis with Sparse Geometries}

\author{Victor Rong$^{*1,2}$
\space\space
Jan Held$^{*3,4}$
\space\space
Victor Chu$^{1,2}$
\space\space
Daniel Rebain$^{5}$
\space\space
Marc Van Droogenbroeck$^{4}$
\\
Kiriakos N. Kutulakos$^{1,2}$
\space\space
Andrea Tagliasacchi$^{\dagger 1,3}$
\space\space
David B. Lindell$^{\dagger 1,2}$\\
\small{\textnormal{$^{1}$University of Toronto\space\space$^{2}$Vector Institute\space\space$^{3}$Simon Frasier University\space\space$^{4}$University of Li\`ege\space\space$^{5}$University of British Columbia}}\\
\small{\textnormal{{$^{*,\dagger}$equal contribution}}}\\
\url{https://lessvrong.com/cs/nexels}
}

\begin{document}

\twocolumn[{
\maketitle
 \vspace{-3em}
 \input{fig/teaser}
}]

\begin{abstract}
Though Gaussian splatting has achieved impressive results in novel view synthesis, it requires millions of primitives to model highly textured scenes, even when the geometry of the scene is simple. We propose a representation that goes beyond point-based rendering and decouples geometry and appearance in order to achieve a compact representation. We use surfels for geometry and a combination of a global neural field and per-primitive colours for appearance. The neural field textures a fixed number of primitives for each pixel, ensuring that the added compute is low. Our representation matches the perceptual quality of 3D Gaussian splatting while using $9.7\times$ fewer primitives and $5.5\times$ less memory on outdoor scenes and using $31\times$ fewer primitives and $3.7\times$ less memory on indoor scenes. Our representation also renders twice as fast as existing textured primitives while improving upon their visual quality.
\end{abstract}
    
\section{Introduction}

In computer graphics, a scene’s geometry governs how light is blocked, while its appearance determines the color contributed to each rendered pixel. Point-based representations like 3D Gaussian splats (3DGS) \textit{merge} these roles, since each primitive jointly encodes geometry and appearance~\cite{Kerbl20233DGaussian}. This coupling stands in contrast to traditional surface-based methods, where coarse geometry (e.g., a quad) can still capture fine-grained appearance (e.g., a detailed painting) through texturing. As illustrated in \Cref{fig:allocation}, Gaussian splats therefore require a large number of primitives to represent high-frequency appearance, leading to substantial memory usage.

A more compact representation becomes possible by \textit{decoupling} geometry from appearance. When appearance is delegated to textures, mesh-based models can depict highly realistic scenes, as commonly seen in professionally produced games and films. Yet meshes have lagged behind in the task of novel view synthesis, where the objective is to generate new viewpoints from a collection of input images of a 3D scene.
In the dense-pose setting we consider, the dominant strategy is \textit{differentiable rendering}, which optimizes scene parameters (geometry, light, and materials) so that rendered views match the training images.
Within this framework, meshes face difficulties due to their discrete connectivity, potential for self-intersections, and complications in computing silhouette gradients~\cite{nicolet2021large, liu2019soft, li2018differentiable, Laine2020Modular}.

\input{fig/allocation}

To mitigate the need for an excessive number of primitives without resorting to the complexities of mesh optimization, recent work has proposed decoupling geometry and appearance in Gaussian splatting by introducing \textit{textured splats}, most commonly via per-primitive image textures~\cite{Rong2025GStex, Chao2025Textured, Svitov2025Billboard,papantonakis2025content}.
However, the memory footprint of such textures grows quadratically with resolution, which constrains their ability to represent high-frequency appearance in large-scale scenes.
\citet{Zhang2025Neural} attempt to overcome this limitation by employing a neural field to provide appearance in a deferred manner: each primitive samples neural features from a multi-resolution hash grid, which are then alpha-composited and passed through a feed-forward network.
While this design requires only a single network evaluation, hash-grid lookups are similarly expensive, and performing several of them for every ray–primitive intersection results in \textit{slow rendering} (${<}30$ FPS), as well as \textit{slow training}.

Although these methods take inspiration from textured meshes, they miss a crucial observation: in standard mesh rasterization with depth buffering, texture fetches are performed only for a \textit{limited} set of \textit{opaque} surface fragments~\cite{Hughes2014CGPP, AkenineMoller2018RealTimeRendering}.
Motivated by this, we introduce a new representation that preserves the optimization advantages of point-based approaches while leveraging the lightweight appearance modeling of mesh-based methods -- all without sacrificing real-time rendering performance.

\paragraph{Contributions.}
We introduce \textit{nexels}, a novel neurally-textured primitive that overcomes the coupling of geometry and appearance in Gaussian splatting.
In particular:
\begin{enumerate*}[label=(\roman*)]
\item For geometry, we introduce a differentiable \textit{quad} indicator, enabling us to better reconstruct surfaces and sharp boundaries;
\item For appearance, we learn a world-space neural field that provides view-dependent texture only for the most relevant primitives, capturing fine details efficiently while keeping computation low;
\item We introduce a new dataset that highlights the limitations of point-based representations in accurately reconstructing regions with high-frequency textures;
\item We achieve perceptual parity with 3DGS using $9.7\times$ fewer primitives on outdoor scenes and $31\times$ fewer primitives on indoor scenes; 
\item Compared to concurrent texture methods, we achieve better photometric quality, while rendering more than twice as fast.
\end{enumerate*}

\section{Related Work}

\subsection{Neural Fields}
A neural field implicitly represents a quantity over a region, such as 3D space, through neural network queries. In novel view synthesis, \citet{Mildenhall2021NeRF} use neural fields to parameterize per-point radiance and density for volumetric rendering. 
These neural radiance fields (NeRFs) have slow rendering speed due to both the large number of samples needed for volumetric rendering as well as expensive per-sample calculations.
To reduce the number of samples used for rendering volumes, early NeRF papers employ stratified sampling and auxiliary proposal networks to improve the sampling efficiency~\cite{Mildenhall2021NeRF, Barron2021MipNeRF, Barron2022MipNeRF360}.
Later works use empty space skipping and other acceleration techniques to avoid unnecessary queries~\cite{Muller2022Instant, Li2023NerfAcc, Sharp2022Spelunking}.
The number of queries can be further constrained using explicit representations such as shells and meshes~\cite{wang2023adaptive, Sharma2024Volumetric}.
In parallel, follow-up methods reduce the computational cost of each query by moving much of the neural network's representational power into spatial features backed by voxel grids~\cite{FridovichKeil2022Plenoxels, Sun2022Direct}, hierarchical grids~\cite{Martel2021Acorn, Takikawa2021Neural, Muller2022Instant, Yu2021PlenOctrees}, point sets~\cite{Chen2023NeuRBF, Xu2022PointNeRF, Chen20243DReconstruction}, and other structures~\cite{Cao2023HexPlane, Chen2022TensoRF}.
Another line of work avoids the slow runtime of neural fields by only using them in an initial stage of training before baking in the appearance and geometry into a fast mesh representation~\cite{Wang2023Neural, Yariv2023BakedSDF, Reiser2024Binary, Guo2023VMesh}.
In contrast, we use surfels to obtain sample locations which avoids the added complication of mesh extraction while seamlessly integrating into the differentiable rendering scheme.

\subsection{Gaussian Splatting}
Gaussian splatting completely forgoes the neural network and instead represents radiance fields as a set of explicit primitives which have decaying opacities according to a Gaussian distribution~\cite{Kerbl20233DGaussian}. It leverages fast point-based rasterization schemes to achieve real-time performance. There has been a deluge of follow-up works, of which we will only focus on those that present new geometry or appearance representations. Several works find that using flattened 3D primitives also results in high-quality novel view synthesis results~\cite{Huang20242DGaussian,Dai2024Highquality} and \citet{Ye2025When} use these to explicitly model opaque surfaces. Other non-Gaussian geometries have also been found to be effective: beta kernels, polynomial kernels, triangles, and more have been proposed~\cite{Held2025Triangle-arxiv,Held2025Triangle+-arxiv,Held20253DConvex,Hamdi2024GES,Liu2025Deformable,MoenneLoccoz20243DGaussian}. Beta splats have been particularly effective at achieving higher rendering quality with fewer parameters. They achieve this reduction through a leaner set of parameters for view-dependent appearance, which is orthogonal to our own work~\cite{Liu2025Deformable}. A number of Gaussian splatting works use neural fields in various ways, but unlike our work, the splats in these works have no spatial variation across them and cannot represent details smaller than themselves ~\cite{Lu2024ScaffoldGS,Ren2025OctreeGS}. 

A recent line of work applies textures to Gaussian splats to increase the appearance capacity of the representation. This has been done through per-primitive image textures~\cite{Rong2025GStex,Chao2025Textured,Svitov2025Billboard,papantonakis2025content} and neural fields~\cite{Xu2024TextureGS,Zhang2025Neural}. Despite promising results, there are three key issues which hinder these methods, summarized in \cref{tab:prior}. First, the majority of these works initialize from a trained dense Gaussian splatting model, which limits their ability to optimize for sparse geometries. Furthermore, the texturing in all prior methods is applied for every ray-primitive interaction. For BBSplat and Textured Gaussians, whose alpha textures further increase overdraw, renders of scene-level reconstructions typically take over fifty milliseconds. Finally, an ideal texture should be able to capture multiscale details within a reasonable amount of memory. Works which use per-primitive image textures are limited by their resolution, which scales poorly with memory. As a partial solution, concurrent work from Papantonakis et al.~\cite{papantonakis2025content} use image textures with adaptive resolutions. Only NeST-Splatting fully overcomes the issue by using a neural field, but this results in much slower renders. Our representation uses a neural field at a fixed number of samples per-pixel, leading to real-time renders.

\begin{table}
    \centering
    \small
    \begin{tabular}{p{1.4in}cccc}
        \toprule
        Method & Scene & Ada & Speed & Mem\\
        \midrule
        Texture-GS~\cite{Xu2024TextureGS} & \faCrossed  & \faCrossed & \faCrossed & \faCrossed\\
        GStex~\cite{Rong2025GStex} & \faChecked & \faCrossed & \faChecked & \faCrossed\\
        Textured Gaussians~\cite{Chao2025Textured} & \faChecked & \faCrossed & \faCrossed & \faCrossed\\
        BBSplat~\cite{Svitov2025Billboard} & \faChecked & \faChecked & \faCrossed & \faCrossed\\
        NeST-Splatting~\cite{Zhang2025Neural} & \faChecked & \faCrossed & \faCrossed & \faChecked\\
        \midrule
        GS-Texturing~\cite{papantonakis2025content} & \faChecked & \faChecked & \faChecked & \faCrossed\\
        Nexels & \faChecked & \faChecked & \faChecked & \faChecked\\
        \bottomrule
    \end{tabular}
    \caption{\textbf{Prior and concurrent textured works.} We summarize the limitations of recent representations with textured primitives. The criteria are \textit{scene-level reconstructions}, \textit{adaptive density control with textures}, \textit{real-time rendering speeds (30+ FPS)}, and \textit{ability to capture texture details in a memory-efficient manner}.
    }
    \label{tab:prior}
\end{table}

\section{Preliminaries}
Our work builds on both Gaussian splatting and neural field methods. In particular, we make use of the differentiable surfel rasterization of 2D Gaussian splatting (2DGS)~\cite{Huang20242DGaussian} and the neural field architecture of Instant-NGP~\cite{Muller2022Instant}.

\paragraph{2D Gaussian Splatting.} Each surfel in 2D Gaussian splatting (2DGS) is a 2D Gaussian in 3D space parameterized by its opacity $\opacity\in \mathbb{R},$ rotation matrix $\rotation = [\ra, \rb, \rc] \in \mathbb{SO}_3,$ mean $\position \in \mathbb{R}^3$, and scale $\scale \in \mathbb{R}^2$. The first two axes of the rotation, $\ra$ and $\rb$, determine the axes of the Gaussian, while $\rc$ functions as the surfel normal. Additionally, each primitive has a view-dependent radiance $\radianceplain \in \mathbb{R}^{3S}$ parameterized by channel-wise spherical harmonics coefficients, where $S$ is the number of coefficients for a predetermined degree.

Following 3DGS, the surfels are preprocessed and sorted by the depth of their centers. For a ray $\ray := \origin + \rayparam \dir$ emanating from the camera position, the intersection in primitive space $(u, v)$ can be computed differentiably from its geometry parameters~\cite{Huang20242DGaussian,Sigg2006GPUBased}. This is used to compute its alpha value, $\alpha := \opacity \exp\left(-(u^2+v^2)/2\right)$. The primitives are alpha-composited in a tile-based rasterizer. The final color is
\begin{equation}
    \radiance(\ray) := \sum_{i} \transmittance_i \alpha_i \radianceplain_i(\ray), \quad \transmittance_i := \prod_{j}^{i-1} (1 - \alpha_i).
    \label{eq:composite}
\end{equation}

\paragraph{Instant-NGP.} Instant-NGP is a neural field architecture composed of a multiresolution hash-grid $\mathcal{H}$ and a tiny multi-layer perceptron (MLP) network $\mathcal{G}$. The multiresolution hash-grid has $L$ levels, each defining a hash-table $\mathcal{H}_\ell$ at given scale $s_\ell$. In our implementation, each hash-table has exactly $T$ entries. Finally, it has a feature dimension of $F$. Altogether it has $L \times T \times F$ parameters.

\begin{figure}
    \centering
    \vspace{-10pt}
    \begin{tikzpicture}[scale=1.0]
        \node[inner sep=0pt] (m) at (0,0)
    {\includegraphics[width=0.99\linewidth]{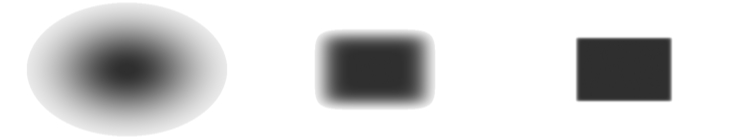}};
        \node[align=center,font=\small] (a) at (-2.7, -1.2) {Diffuse};
        \node[align=center,font=\small] (b) at (2.8, -1.2) {Sharp};
        \draw [->,line width=1pt] (a) -- (b);
    \end{tikzpicture}
    \vspace{-5pt}
    \caption{\textbf{Kernel.} We use a generalized Gaussian kernel in order to model near-opaque primitives. Values below $0.1$ are set to $0$ for visualization purposes. (Left) For $\gamma {=} 1$, our nexels correspond to a Gaussian distribution. (Right) As $\gamma {\rightarrow} \infty$, they converge toward the indicator of a quad.
    }
    \label{fig:kernel}
\end{figure}

Given an input coordinate $x \in \mathbb{R}^3$, an Instant-NGP $\mathcal{F}$ can be configured to return a vector whose size is the MLP output dimension. First, a hash $\mathbf{h}_\ell \in \{0, 1, \ldots, T-1\}$ is calculated for each hash-grid level $\ell$ based on $s_\ell \cdot x$. This hash is indexed into the grid at that level to retrieve a feature vector. These are concatenated to obtain \begin{equation}
    \mathcal{H}(x) := \left(\mathcal{H}_{0,\mathbf{h}_0}, \ldots, \mathcal{H}_{L-1,\mathbf{h}_{L-1}}\right) \in \mathbb{R}^{L \times F}.
\end{equation} Finally, the MLP is applied to yield $\mathcal{F}(x) := \mathcal{G}(\mathcal{H}(x)).$

\begin{figure*}
    \centering
    \footnotesize
    \begin{overpic}[width=1.0\linewidth]{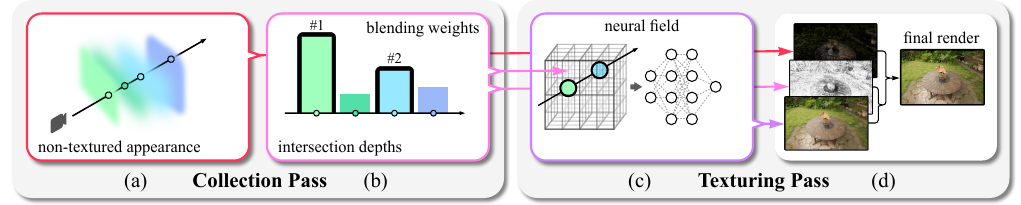}
    \put(116,170){Eq.~\ref{eq:notexture}}
    \put(82,72){$\radianceplain_0$}
    \put(112,72){$\radianceplain_1$}
    \put(142,72){$\radianceplain_2$}
    \put(172,72){$\radianceplain_3$}
    \put(223,140){\normalsize{$\Notextures$}}
    
    \put(300,97){$\weight_0$}
    \put(338,97){$\weight_1$}
    \put(376,97){$\weight_2$}
    \put(414,97){$\weight_3$}
    \put(302,70){$t_0^*$}
    \put(340,70){$t_1^*$}
    \put(378,70){$t_2^*$}
    \put(416,70){$t_3^*$}
    \put(458,125){\normalsize{$\Depths$}}
    \put(455,106){\normalsize{$\Weights$}}

    \put(560,58){$\mathcal{H}$}
    \put(661,58){$\mathcal{G}$}
    \put(610,52){Eq.~\ref{eq:texture}}
    \put(710,111){$\radiancetexture$}
    \put(718,73){\normalsize{$\Textures$}}

    \put(907,74){Eq.~\ref{eq:final}}
    \end{overpic}
    \caption{\textbf{Rendering pipeline.} (a) We iterate over the surfels and determine the weights at each intersection while also returning a non-textured render. (b) We collect the $K$ intersections which have the largest weights. (c) We pass their $K$ positions into a neural field to obtain textured appearance. (d) We composite the textures with the non-textured render to get the final output.
    }
    \label{fig:method}
\end{figure*}

\section{Nexels}

This section provides an overview of the main components of our method. \Cref{sec:representation} introduces the geometry and appearance representations, while \Cref{sec:rendering} describes the rendering process. Finally, \Cref{sec:optimization} outlines the optimization framework, including adaptive density control and the loss terms used during training. For notational convenience, we present our method for a single pixel $p$ with ray $\ray$ and use view-\textit{independent} radiance in equations.

\subsection{Representation}\label{sec:representation}

\paragraph{Geometry.}
We represent the geometry of nexels using a set of $N$ surfels defined by a kernel governing their opacities. Following 2DGS, the $i^{\text{th}}$ primitive has a mean $\mu_i\in \mathbb{R}^3$, rotation $\rotation_i \in \mathbb{SO}_3$, scale $\sigma_i \in \mathbb{R}^2$, and opacity $\opacity_i\in \mathbb{R}$. In addition, we have a gamma parameter $\gamma_i \in \mathbb{R}^2$ that controls how rect-like the primitive is.
The opacity at an intersection $(u, v)$ in primitive space is defined by the generalized Gaussians,
\begin{equation}
    \alpha := \opacity_i \exp\left(-\frac{|u|^{2\gamma_{i,1}}}{2}\right)\exp\left(-\frac{|v|^{2\gamma_{i,2}}}{2}\right).
\end{equation}
\Cref{fig:kernel} shows the nexel shape for different values of $\gamma$.
As~$\gamma {\rightarrow} 1$, the kernel converges to a standard Gaussian.
As~$\gamma {\rightarrow} \infty$ it converges toward a quad with a sharp transition. 
This kernel is well suited for modeling opaque and flat surfaces: for large $\opacity$ and $\gamma$, the function remains close to $1$ over most of the support, unlike a conventional Gaussian which has a long tail.
Our kernel behaves like a Gaussian where smoothness is needed, while still being able to produce hard edges in regions with sharp transitions.
We constrain $\gamma {\ge} 1$, which guarantees that our kernel is $C_{\infty}$.
Taking inspiration from the separable filters of signal processing, we choose to take the outer product between the $u$ and $v$ axes. As $\gamma$ grows, the kernel approaches a rectangle rather than an ellipse as in prior work~\cite{Hamdi2024GES,Liu2025Deformable}.

\paragraph{Appearance.}
Each surfel has two modes of appearance: non-textured and textured.
During rendering, we select which to use on a per-pixel basis.
The non-textured appearance of surfel $i$ is the same as 2DGS~\cite{Huang20242DGaussian}, $\radianceplain_i \in \mathbb{R}^{3S}$.

The textured appearance of each primitive is represented using a shared neural field $\mathcal{F}$ backed by an Instant-NGP architecture~\cite{Muller2022Instant}. The field takes as input any 3D position $x \in \mathbb{R}^3$ and outputs a view-dependent radiance $\mathcal{F}(x)$ as a set of spherical harmonics coefficients. The texture at a given point on any of the surfels can be computed by querying the neural field at that position.

For a primitive $i$, we define $\rayparam^*_i(\ray)$ as the intersection depth and $x^*_i(\ray) := \origin + \rayparam^* \dir$ as the world-space intersection. Naively, the textured appearance for pixel $p$ would be $\mathcal{F}(x^*_i(\ray)).$
Using point samples for differentiable rendering leads to inaccuracies in the fine levels of the hash-grid.
Following the down-weighting analysis of \citet{Barron2023ZipNeRF}, we multiply the hash-grid features of level $\ell$ by
\begin{equation}\label{eq:downweighting}
    \Delta_{i,\ell}(\ray) := 1 - \exp\left(-\frac{1}{2\pi} \left(\frac{f}{s_\ell t_i^*(\ray)}\right)^2\right),
\end{equation} where $f$ is the focal length. This requires the same amount of computation as the naive sampling and removes noticeable aliasing artifacts. Ultimately, we define the filtered texture radiance along $\ray$ for primitive $i$ as \begin{equation} \label{eq:texture}
\radiancetexture_i(\ray) :=\mathcal{G}\left(\Delta_i(\ray) \cdot \mathcal{H}(x_i^*(\ray))\right).
\end{equation}
Note that the only primitive information needed to compute $\radiancetexture_i(\ray)$ is the intersection depth $t_i^*(\ray)$.

\subsection{Rendering}
\label{sec:rendering}

Our rendering step is composed of \textit{two passes} over the image pixels. The first pass computes the initial render using only the non-textured appearance and collects information on which primitives should be textured. The second pass applies the neural texture at the collected ray-primitive intersections and updates the initial render.
\Cref{fig:method} provides an overview of the different rendering passes.

\paragraph{Collection Pass.}

In the first pass, we compute the alpha $\alpha_i$ and transmittance values $\transmittance_i$ across the $M$ surfels which intersect a pixel's ray. We use \cref{eq:composite} to compute the initial render $\Notextures$ for each pixel, assuming no primitive is textured.

When rendering a single pixel, we limit the number of primitives which are textured to a hyperparameter~$K {\ll} M$. We determine the subset of primitives which are textured based on their blending weights which we denote as $\weight_i := \alpha_i \transmittance_i$. For a given pixel, the $K$ primitives with the highest weights are selected. Inspired by fragment buffer techniques~\cite{Bavoil2007Multifragment}, we maintain a per-pixel buffer of size $K$. Each buffer entry contains a primitive id, its weight, and its depth, totalling to $3K$ 32-bit registers. Initially, the buffer is empty. When iterating through the $M$ primitives and computing the weights, we update the buffer to store the running primitives with the highest weights. The final id, depth, and weight buffers are written into $H \times W \times K$ images $\Ids, \Depths,$ and $\Weights$, respectively.  These are returned along with the non-textured render $\Notextures$. Additionally, for the $K$ primitives in $\Ids$, we remove their non-textured colour from the render $\Notextures$. The final value of the non-textured render at pixel $p$ is \begin{equation}\label{eq:notexture}
    \Notextures_p := \sum_{i=0}^{M-1} \left[i \not\in \Ids_p\right]\weight_i\radianceplain_i.
\end{equation}
\paragraph{Texturing Pass.}
The texturing pass first computes world-space intersection positions from the depth-map buffer $\Depths$, yielding $x \in \mathbb{R}^{H\times W\times K\times 3}$. Following \cref{eq:texture}, we query the neural field to compute the filtered texture $\Textures \in \mathbb{R}^{H\times W\times K\times 3}$ where $\Textures_{p,j} := \radiancetexture_{\Ids_{p,j}}(\ray)$ for pixel $p$ with corresponding ray $\ray$. Extending \cref{eq:composite} to include the choice between texturing and not texturing, we have
\begin{align}
    \sum_{i=0}^{M-1} \transmittance_i \alpha_i \radiance_i(\ray) &= \sum_{i=0}^{M-1} \weight_i \left(\left[i \not\in \Ids_p\right]\radianceplain_i + \left[i \in \Ids_p\right]\radiancetexture_i(\ray)\right)\nonumber\\
    &= \Notextures_p + \sum_{j=0}^{K-1} \Weights_{p,j}\Textures_{p,j}. \label{eq:final}
\end{align}
Hence, we compute the final render from $\Notextures, \Weights,$ and $\Textures$.

\subsection{Optimization}\label{sec:optimization}

We follow the scheme of 3DGS~\cite{Kerbl20233DGaussian} and optimize against training views for $30{,}000$ iterations with regularizations. We also interleave density control operations.

\paragraph{Adaptive Density Control.}
We introduce a hyperparameter $P$ to control the number of primitives. Similar to BBSplat~\cite{Svitov2025Billboard}, we sample the point cloud output of COLMAP for initialization~\cite{Schonberger2016Structure}.
Specifically, we use farthest point sampling to reduce the initial point cloud to $0.5P$ points~\cite{Li2022AnAdjustable}. We initialize the scale at each point based on its distance to its predecessors during farthest point sampling.

We also prune unnecessary primitives and add new ones in regions where the current geometry is insufficient. We perform a densification and pruning step every $100$ iterations from iteration $500$ to $25{,}000$. We select $5\%$ of primitives to split evenly along their longest axis.
The selection is stochastic with probabilities defined using blended errors similar to \citet{RotaBulo2024Revising}'s method. We cap the number of primitives added in the splitting by $P$. Afterward, we prune all primitives whose opacity is below $0.005$.

\paragraph{Losses.} At each iteration, we render the nexels from a given training view to produce a predicted image. Following \citet{Kerbl20233DGaussian}, we compute a photometric loss between the prediction and ground truth using a mixture of an $L_1$ loss $\mathcal{L}_1$ and D-SSIM loss $\mathcal{L}_{\text{D-SSIM}}$~\cite{Kerbl20233DGaussian},
\begin{equation}
    \mathcal{L}_{\text{image}} := (1 - \lambda_{\text{D-SSIM}})\mathcal{L}_1 + \lambda_{\text{D-SSIM}} \mathcal{L}_{\text{D-SSIM}}.
\end{equation}
We also include a texture loss, $\mathcal{L}_{\text{texture}},$ so that the non-premultiplied texture fits to the ground truth image $I^{\text{GT}}$: \begin{equation}
    \frac{1}{3HW}\sum_{y=0}^{H-1}\sum_{x=0}^{W-1}\sum_{c=0}^{2} \left|I^{\text{GT}}_{y,x,c} - \frac{\sum_{j=0}^{K-1} \Weights_{y,x,j}\Textures_{y,x,j,c}}{\sum_{j=0}^{K-1} \Weights_{y,x,j}}\right|.
\end{equation}
To penalize non-textured rendering, we minimize the total blending weights of the non-textured appearance,
\begin{equation}
    \mathcal{L}_{\text{alpha}} := \frac{1}{HW} \sum_{y=0}^{H-1}\sum_{x=0}^{W-1}\left(1-\sum_{j=0}^{K-1} \Weights_{y,x,j}\right).
\end{equation}
We include an $L_1$ loss on the opacity to induce sparsity~\cite{Kheradmand20243DGaussian},
\begin{equation}
    \mathcal{L}_{\text{opacity}} := \frac{1}{N} \sum_{i=0}^{N-1}\opacity_i.
\end{equation}
Finally, we add a grid weight regularization term,
\begin{equation}
    \mathcal{L}_{\text{grid}} := \sum_{\ell=0}^{L-1} \sum_{i=0}^{T-1} \sum_{j=0}^{F-1} s_i^{-3} \mathcal{H}_{\ell,i,j}
\end{equation}
which encourages zero-mean grid values that justify the down-weighting done for grid anti-aliasing~\cite{Barron2023ZipNeRF}.
The final loss is
\begin{alignat}{2}
    \mathcal{L} := \mathcal{L}_{\text{image}}
    &+\lambda_{\text{alpha}}\mathcal{L}_{\text{alpha}} &&+\lambda_{\text{texture}}\mathcal{L}_{\text{texture}} \nonumber\\
    &+\lambda_{\text{opacity}}\mathcal{L}_{\text{opacity}} &&+ \lambda_{\text{grid}}\mathcal{L}_{\text{grid}}.
\end{alignat}
We set  $\lambda_{\text{alpha}} = 0.005, \lambda_{\text{texture}} = 0.5, \lambda_{\text{opacity}} = 0.01,\lambda_{\text{grid}} = 0.01,$ and $\lambda_{\text{D-SSIM}}=0.2$ for all experiments.

\begin{figure*}
    \centering
    \includegraphics[width=0.99\linewidth]{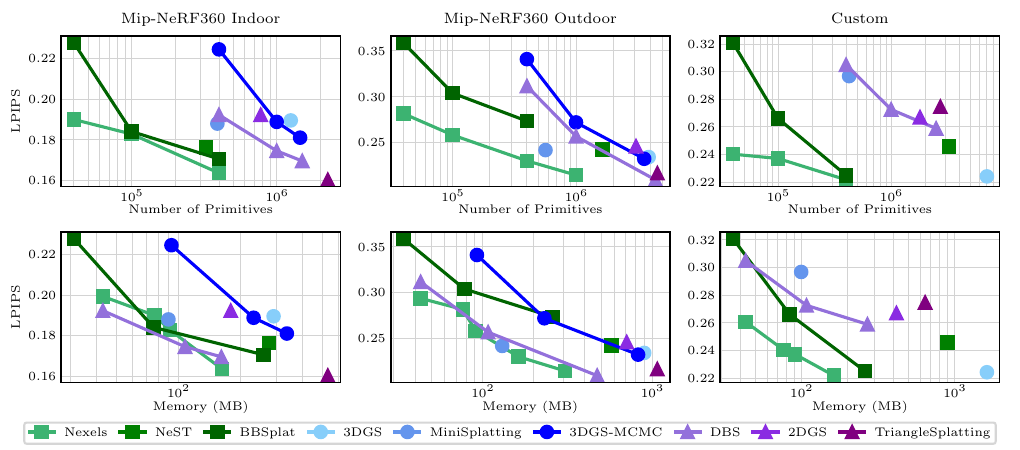}
    \caption{\textbf{Quality versus representation size.} We show how the perceptual quality of our representation varies depending on the number of primitives (top row) and memory (bottom row). We evaluate the LPIPS$\downarrow$ across multiple settings for several methods and display separate graphs averaged over the four Mip-NeRF360 indoor scenes (left column), five Mip-NeRF360 outdoor scenes (middle column), and the four highly textured scenes of our custom dataset (right column). Nexels consistently has the lowest LPIPS for any number of primitives under $10^6$, for all three of the datasets. When plotted against memory, nexels is always in the top three at any amount.
    }
    \label{fig:chart_number}
\end{figure*}

\begin{table*}[t!]
	\small
\resizebox{\linewidth}{!}{
  \setlength{\tabcolsep}{4pt}
\begin{tabular}{lccccccccccccccc}
    \toprule
	& \multicolumn{5}{c}{Mip-NeRF360} & \multicolumn{5}{c}{Tanks \& Temples} & \multicolumn{5}{c}{Custom}\\
	& PSNR & SSIM & LPIPS & Memory & \#Prim.
	& PSNR & SSIM & LPIPS & Memory & \#Prim.
    & PSNR & SSIM & LPIPS & Memory & \#Prim.\\
    \cmidrule(l{3pt}r{3pt}){1-1}\cmidrule(l{3pt}r{3pt}){2-6}\cmidrule(l{3pt}r{3pt}){7-11}\cmidrule(l{3pt}r{3pt}){12-16}
    3DGS \cite{Kerbl20233DGaussian} & 27.21 & 0.815 & 0.214 & 640MB & 2.7M & 23.80 & 0.853 & 0.169 & 348MB & 1.5M & 26.65 & \textbf{0.858} & 0.224 & 1630MB & 7.0M \\
    2DGS \cite{Huang20242DGaussian} & 27.04 & 0.805 & 0.252 & 480MB & 2.0M & 23.15 & 0.831 & 0.212 & 200MB & 0.8M & 25.98 & 0.831 & 0.267 & 450MB & 1.8M \\
    DBS \cite{Liu2025Deformable} & \textbf{28.10} & \textbf{0.829} & 0.192 & 340MB & 3.1M & \textbf{24.82} & \textbf{0.871} & 0.144 & 260MB & 2.0M & 26.21 & 0.842 & 0.259 & 270MB & 2.5M \\
    TriSplat \cite{Held2025Triangle-arxiv} & 27.16 & 0.814 & \textbf{0.191} & 790MB & 3.5M & 23.14 & 0.857 & \textbf{0.143} & 400MB & 2.0M & 24.28 & 0.780 & 0.275 & 620MB & 2.7M \\
    \cmidrule(l{3pt}r{3pt}){1-1}\cmidrule(l{3pt}r{3pt}){2-6}\cmidrule(l{3pt}r{3pt}){7-11}\cmidrule(l{3pt}r{3pt}){12-16}
    NeST-Splat \cite{Zhang2025Neural} & 26.68 & 0.795 & 0.212 & 440MB & 1.0M & 23.02 & 0.824 & 0.181 & 260MB & 0.5M & 25.40 & 0.824 & 0.246 & 900MB & 3.2M \\
    BBSplat \cite{Svitov2025Billboard} & 26.98 & 0.783 & 0.231 & 260MB & 0.4M & 23.83 & 0.854 & 0.147 & 260MB & 0.4M & 26.57 & 0.844 & 0.225 & 260MB & 0.4M \\
    Ours & 27.35 & 0.802 & 0.201 & 160MB  & 0.4M & 23.55 & 0.841 & 0.155 & 160MB & 0.4M & \textbf{27.04} & 0.851 & \textbf{0.222} & 160MB & 0.4M \\
    \bottomrule
\end{tabular}
	}
\caption{\textbf{Quantitative results.} We compare nexels with 400K primitives to baselines at their maximum recommended number of primitives, corresponding to each baseline's rightmost point on \Cref{fig:chart_number}.
}
\label{tab:high_results}
\end{table*}

\input{fig/low_number}

\input{fig/low_memory}

\section{Results}
Our method is implemented in Python and CUDA on top of the INRIA 3DGS codebase~\cite{Kerbl20233DGaussian}.
We use PyTorch for performing parameter activations and composing the render passes with autograd and use the Adam optimizer for all parameters~\cite{Paszke2019Pytorch, Kingma2014Adam}.
We implement custom CUDA kernels for our multiresolution hash-grid operations and rasterization passes.
We use the NVIDIA tinycudann library for the fully-fused MLP pass~\cite{Muller2022Instant}.
For all scenes, we use a texture limit of just $K=2$.
The MLP has two hidden layers without bias terms, each with a width of $64$, and ReLU activations.
We use degree $3$ spherical harmonics coefficients for both appearance modes, corresponding to $S=16$ and an MLP output dimension of $48$.
For the hash-grid, $L=16$ and $F=2$.
Unless otherwise stated, we use $T=2^{20}$.

\subsection{Datasets and Metrics}
We perform our evaluation over fifteen real-world scenes across three datasets: all nine scenes from Mip-NeRF360~\cite{Barron2022MipNeRF360}, the \textsc{Train} and \textsc{Truck} scenes from the Tanks \& Temples dataset~\cite{Knapitsch2017Tanks}, and four scenes which we collect involving high amounts of texture detail. Existing datasets seldom have close-up views despite these being important for photorealism. Fine appearance details within these scenes barely show up in their metrics. Our scenes contain high-frequency appearance, such as text and patterns, observed at a range of distances, including close-up shots. We describe the capture and preprocessing details in the supplementary. The full dataset will be released publicly.

We compare against point-based representations~\cite{Kerbl20233DGaussian, Huang20242DGaussian, Held2025Triangle-arxiv, Liu2025Deformable} and textured representations~\cite{Svitov2025Billboard, Zhang2025Neural}, which are state-of-the-art for novel view synthesis. We refer to the supplementary for experimental details for our baselines.
In addition, we consider Gaussian splat variants which can produce sparser point sets than the original 3DGS~\cite{Fang2024MiniSplatting, Kheradmand20243DGaussian}.

We evaluate photometric quality with the standard PSNR, SSIM, and LPIPS metrics used in novel view synthesis and differentiable rendering~\cite{Mildenhall2021NeRF,Zhang2018TheUnreasonable}. The LPIPS metric across all baselines is computed with the same normalization as the original 3DGS paper~\cite{Kerbl20233DGaussian,RotaBulo2024Revising}. All timings are performed on a single RTX 6000 Ada GPU using the CUDA implementation of each method's forward pass.

\subsection{Evaluation}

\paragraph{Novel-view synthesis.} We compare all baselines at each of their maximum recommended settings in \cref{tab:high_results}. On the Mip-NeRF360 dataset, nexels are competitive with point-based representations with only 400K primitives, achieving better PSNR than all methods except Deformable Beta Splatting (DBS~\cite{Liu2025Deformable}), which has over $7\times$ the number of primitives and twice the memory. On our custom dataset, we achieve the highest PSNR and LPIPS and come second in SSIM to 3DGS at $10\times$ our memory. We achieve better scores than NeST-Splatting and BBSplat on the Mip-NeRF360 and custom datasets.

\paragraph{Visual quality under primitive and memory budgets.}
In ~\Cref{fig:chart_number}, we plot the perceptual quality (LPIPS) against the number of primitives and memory across three scene groups. For baselines which can specify a number of primitives, we include data points corresponding to appropriate subsets of $\{40\text{K}, 100\text{K}, 400\text{K}, 1\text{M}\}$ primitives. Details are included in the supplementary material.

Our representation exceeds virtually all other models at equal numbers of primitives. For example, 400K nexels obtain an LPIPS score of $0.164$ across Mip-NeRF360 indoor scenes, while BBSplat and DBS obtain LPIPS scores of $0.171$ and $0.192$, respectively. Conversely, to reach a given LPIPS, nexels require far fewer primitives than other methods. The best result of 3DGS on the Mip-NeRF360 outdoor scenes is $0.234$ LPIPS with 3.9 million Gaussians. We achieve $0.230$ LPIPS with just 400K nexels. Similarly, 3DGS obtains $0.190$ LPIPS on indoor scenes with 1.2 million Gaussians. We reach the same score at 40K nexels. Our method is also competitive when measured against memory. At 40MB on MipNeRF-360 outdoor scenes, we achieve $0.294$ LPIPS while DBS achieves $0.312$ LPIPS. We include qualitative results echoing these trends in Figures \ref{fig:low_number} and \ref{fig:low_memory}.

For the custom scenes with high amounts of texture, nexels are clearly better than non-textured representations. We achieve an LPIPS of $0.240$ with just 40K nexels, outperforming DBS with $2.5$ million splats, which obtains $0.259$ LPIPS. BBSplat comes near us at 400K primitives but struggles at low primitive counts.

\begin{table}
    \centering
    \small
    \begin{tabular}{lcccccc}
    \toprule
    & \multicolumn{2}{c}{Indoor} & \multicolumn{2}{c}{Outdoor} & \multicolumn{2}{c}{Mean}\\
    Method & FPS & Train. & FPS & Train. & FPS & Train. \\
    \cmidrule(l{3pt}r{3pt}){1-1}\cmidrule(l{3pt}r{3pt}){2-3}\cmidrule(l{3pt}r{3pt}){4-5}\cmidrule(l{3pt}r{3pt}){6-7}
    NeST-Splat & 27 & 4.4h & 19 & 5.2h & 23 & 4.9h\\
    BBSplat & 20 & 2.4h & 20 & 2.1h & 20 & 2.2h\\
    Nexels & \textbf{40} & \textbf{1.5h} & \textbf{58} & \textbf{0.9h} & \textbf{50} & \textbf{1.1h}\\
    \bottomrule
    \end{tabular}
    \caption{\textbf{Timing comparisons.} We show the average FPS of test-views across the Mip-NeRF360 scenes for the textured representations as well as training times.}
    \label{tab:timing_comparison}
\end{table}

\paragraph{Training time and rendering speed.}
\Cref{tab:timing_comparison} compares training time and rendering speed for different texturing works. We separate between indoor and outdoor scenes as they have different image sizes. NeST-Splat averages five hours of training and only renders at 23 FPS across Mip-NeRF360 scenes.
This is due to its texturing, where hash-grid lookups are done for each ray–primitive intersection, totaling thousands of random accesses per ray.
BBSplat has a faster training time, but their use of alpha textures results in high overdraw, leading to a rendering speed of 20 FPS.
In contrast, nexels train $4.2$ and $1.9$ times faster than NeST-Splatting and BBSplat, respectively, and render at 50 FPS.

\subsection{Ablations}

The results in \cref{tab:ablation} ablate individual components of our method to demonstrate their importance. We evaluate each individual experiment across the Mip-NeRF360 scenes trained with 400K primitives.

First, we verify that the main component, the neural texture, is useful. We find that all visual metrics, especially LPIPS, worsen without the neural texture. Qualitatively, the non-textured renders lose background details. Removing the $\gamma$ parameter and using 2D Gaussians as the kernel results in slight degradations overall as a Gaussian kernel leads to lower texture weights. Dropping the higher-order per-primitive spherical harmonics coefficients and only relying on the neural field for view-dependent effects is a possibility if memory is the main priority, as it removes 45 parameters for each primitive. The increase in LPIPS is very small, although there is a trade-off in PSNR. Finally, omitting the hash-grid downweighting factor in \cref{eq:downweighting} leads to noticeable aliasing artifacts in test views, as the fine levels overfit at the intersections seen from training poses.

\begin{table}
    \centering
	\small
    \begin{tabular}{lccccc}
    \toprule
    & PSNR & SSIM & LPIPS & Mem. & FPS\\
    \midrule
    No texture & 27.20 & 0.787 & 0.261 & 100MB & \textbf{85} \\
    No gamma & 27.10 & 0.797 & 0.210 & 160MB & 48\\
    No prim. SH & 27.00 & 0.797 & 0.205 & \textbf{90MB} & 47\\
    No down. & 27.03 & 0.797 & 0.211 & 160MB & 49\\
    \midrule
    Full & \textbf{27.35} & \textbf{0.802} & \textbf{0.201} & 160MB & 50\\
    \bottomrule
    \end{tabular}
    \caption{\textbf{Ablations.} We show how dropping individual elements of nexels affects the photometric quality, memory, and FPS of the representation averaged across all Mip-NeRF360 scenes.}
    \label{tab:ablation}
\end{table}

\subsection{Limitations}

The speed of the Instant-NGP model is reliant on tensor core operations, which are only available on high-end GPUs. While nexels are real-time on the machinery we use, real-time rendering on mobile GPUs or low-end laptop GPUs is currently not possible. In addition, nexels produce noise in unseen regions, which is likely less appealing for end users than the blurry artifacts of Gaussian splats. Our work also does not support the full complexity of the imaging process, including motion blur and depth-of-field blur. These effects are particularly relevant for differentiable rendering of fine details, as even small deviations from the expected pinhole camera model can damage reconstructions.

\section{Conclusion}

We present the first representation for novel view synthesis that removes the reliance on dense primitives while achieving real-time rendering at high quality. Our results show that far fewer geometries are needed to model scene-level datasets than recent research would suggest. Future work could incorporate level-of-detail structures to model city-scale scenes. Sparse surfaces also work well for ray-tracing applications. Finally, the top-$K$ scheme can be extended to support textures combining the neural field and per-primitive features. In conclusion, nexels break new ground in the exploration between volumetric and surface representations, capturing high-frequency appearance at a low memory and compute cost.

\paragraph{Acknowledgements.} JH acknowledges support from the F.R.S.-FNRS. DBL and KNK acknowledge support from NSERC under the RGPIN program. DBL also acknowledges support from the Canadian Foundation for Innovation and the Ontario Research Fund.

{
    \small
    \bibliographystyle{ieeenat_fullname}
    \bibliography{bib/abbreviation-empty,
    bib/abbreviation-short,bib/all, bib/new_references}

@STRING{publisher-ieee = ""}

@STRING{publisher-ieee-short = ""}

@STRING{publisher-springerswiss = ""}

@STRING{arxiv = "CoRR"}

@STRING{arxiv = "arXiv"}

@STRING{publisher-eurass = "Eurographics Assoc."}

@STRING{acm-cacm = "Commun. ACM"}

@STRING{acm-tog = "ACM Trans. Graph."}

@STRING{acm-c-siggraph-cp = "ACM SIGGRAPH Conf. Proc."}

@STRING{acm-c-siggraph-aisa-papers = "ACM SIGGRAPH Asia Conf. Pap."}

@STRING{acm-c-siggraph-papers = "ACM SIGGRAPH Conf. Pap."}

@STRING{acm-s-igg = "ACM Symp. Interact. 3D Graph. Games"}

@STRING{publisher-ieee = "Inst. Electr. Electron. Eng. (IEEE)"}

@STRING{publisher-ieee-short = "IEEE"}

@STRING{ieee-tpami = "IEEE Trans. Pattern Anal. Mach. Intell."}

@STRING{ieee-c-cvpr = "IEEE Conf. Comput. Vis. Pattern Recognit. (CVPR)"}

@STRING{ieee-c-iccv = "IEEE Int. Conf. Comput. Vis. (ICCV)"}

@STRING{ieee-c-iccvx = "IEEE/CVF Int. Conf. Comput. Vis. (ICCV)"}

@STRING{ieee-c-icvprx = "IEEE/CVF Conf. Comput. Vis. Pattern Recognit. (CVPR)"}

@STRING{ieee-c-wacvx = "IEEE/CVF Winter Conf. Appl. Comput. Vis. (WACV)"}

@STRING{ieee-w-sips = "IEEE Work. Signal Process. Syst. (sips)"}

@STRING{publisher-springerswiss = "Springer Nat. Switz."}

@STRING{as = "Acta Ster."}

@STRING{lncs   = "Lect. Notes Comput. Sci."}

@STRING{eccv = "Eur. Conf. Comput. Vis. (ECCV)"}

@STRING{nips = "Adv. Neural Inf. Process. Syst. (NeurIPS)"}

@STRING{threedv = "Int. Conf. 3D Vis. (3DV)"}

@STRING{publisher-acmlong = "Assoc. Comput. Mach."}

@STRING{publisher-curran = "Curran Assoc. Inc."}

@STRING{city-boston = "Boston, MA, USA"}

@STRING{city-denver = "Denver, CO, USA"}

@STRING{city-honolulu = "Honolulu, HI, USA"}

@STRING{city-lasvegas = "Las Vegas, NV, USA"}

@STRING{city-losangeles = "Los Angeles, CA, USA"}

@STRING{city-montreal = "Montr{\'e}al, Can."}

@STRING{city-nashville = "Nashville, TN, USA"}

@STRING{city-neworleans = "New Orleans, LA, USA"}

@STRING{city-paris = "Paris, Fr."}

@STRING{city-saltlake = "Salt Lake City, UT, USA"}

@STRING{city-seattle = "Seattle, WA, USA"}

@STRING{city-seoul = "Seoul, South Korea"}

@STRING{city-sydney = "Sydney, New South Wales, Aust."}

@STRING{city-telaviv = "Tel Aviv, Isra{\"e}l"}

@STRING{city-tucson = "Tucson, AZ, USA"}

@STRING{city-vancouver = "Vancouver, Can."}

@STRING{Jan = "Jan."}

@STRING{Mar = "Mar."}

@STRING{Apr-May = "Apr.-May"}

@STRING{Jun = "Jun."}

@STRING{Jul = "Jul."}

@STRING{Oct = "Oct."}

@STRING{Nov = "Nov."}

@STRING{Dec = "Dec."}

@article{Laine2020Modular,
	title = {Modular primitives for high-performance differentiable rendering},
	author = {Laine, Samuli and Hellsten, Janne and Karras, Tero and Seol, Yeongho and Lehtinen, Jaakko and Aila, Timo},
	journal = acm-tog,
	volume = {39},
	number = {6},
	pages = {1-14},
	month = Nov,
	year = {2020},
	publisher = publisher-acmlong,
	doi = {10.1145/3414685.3417861},
	url = {https://doi.org/10.1145/3414685.3417861}
}

@inproceedings{Schonberger2016Structure,
        title = {Structure-from-Motion Revisited},
        author = {Schonberger, Johannes L. and Frahm, Jan-Michael},
        booktitle = ieee-c-cvpr,
        shortjournalproceedings = {2016 IEEE Conf. Comput. Vis. Pattern Recognit. (CVPR)},
        pages = {4104-4113},
        month = Jun,
        year = {2016},
        publisher = publisher-ieee,
        address = city-lasvegas,
        doi = {10.1109/cvpr.2016.445},
        url = {https://doi.org/10.1109/CVPR.2016.445}
}

@article{Knapitsch2017Tanks,
        title = {Tanks and temples: benchmarking large-scale scene reconstruction},
        author = {Knapitsch, Arno and Park, Jaesik and Zhou, Qian-Yi and Koltun, Vladlen},
        journal = acm-tog,
        volume = {36},
        number = {4},
        pages = {1-13},
        month = Jul,
        year = {2017},
        publisher = publisher-acmlong,
        doi = {10.1145/3072959.3073599},
        url = {https://doi.org/10.1145/3072959.3073599}
}

@article{Held2025Triangle+-arxiv,
	title = {Triangle Splatting+: Differentiable Rendering with Opaque Triangles},
	author = {Held, Jan and Vandeghen, Renaud and Son, Sanghyun and Rebain, Daniel and Gadelha, Matheus and Zhou, Yi and Lin, Ming C. and Van Droogenbroeck, Marc and Tagliasacchi, Andrea},
	journal = arxiv,
	volume = {abs/2509.25122},
	year = {2025},
	publisher = arxiv,
	eprint = {2509.25122},
	eprinttype = {arXiv},
	doi = {10.48550/arXiv.2509.25122},
	url = {https://doi.org/10.48550/arXiv.2509.25122}
}

@inproceedings{Held20253DConvex,
	title = {{3D} Convex Splatting: Radiance Field Rendering with {3D} Smooth Convexes},
	author = {Held, Jan and Vandeghen, Renaud and Hamdi, Abdullah and Deli{\`e}ge, Adrien and Cioppa, Anthony and Giancola, Silvio and Vedaldi, Andrea and Ghanem, Bernard and Van Droogenbroeck, Marc},
        booktitle = ieee-c-icvprx,
	pages = {21360-21369},
        month = Jun,
        year = {2025},
        address = city-nashville,
	publisher = publisher-ieee,
	doi = {10.1109/CVPR52734.2025.01990},
	url = {https://doi.org/10.1109/CVPR52734.2025.01990}
}

@inproceedings{Held2026Triangle,
	title = {Triangle Splatting for Real-Time Radiance Field Rendering},
	author = {Held, Jan and Vandeghen, Renaud and Deli{\`e}ge, Adrien and Hamdi, Abdullah Rebain, Daniel and Giancola, Silvio and Cioppa, Anthony and Vedaldi, Andrea and Ghanem, Bernard and Tagliasacchi, Andrea and Van Droogenbroeck, Marc},
	booktitle = threedv,
	pages = {1-10},
	month = Mar,
	year = {2026},
	address = city-vancouver,
	doi = {},
	url = {}
}

@article{Held2025Triangle-arxiv,
	title = {Triangle Splatting for Real-Time Radiance Field Rendering},
	author = {Held, Jan and Vandeghen, Renaud and Deli{\`e}ge, Adrien and Hamdi, Abdullah Rebain, Daniel and Giancola, Silvio and Cioppa, Anthony and Vedaldi, Andrea and Ghanem, Bernard and Tagliasacchi, Andrea and Van Droogenbroeck, Marc},
	journal = arxiv,
	volume = {abs/2505.19175},
	year = {2025},
	publisher = arxiv,
	eprint = {2505.19175},
	eprinttype = {arXiv},
	doi = {10.48550/arXiv.2505.19175},
	url = {https://doi.org/10.48550/arXiv.2505.19175}
}

@article{Kingma2014Adam,
	title = {Adam: A Method for Stochastic Optimization},
	author = {Kingma, Diederik and Ba, Jimmy},
	journal = arxiv,
	volume = {abs/1412.6980},
	eprint = {1412.6980},
	url = {http://arxiv.org/abs/1412.6980v9},
	year = 2014,
	month = Dec  
}

@inproceedings{Paszke2019Pytorch,
        title = {Pytorch: An imperative style, high-performance deep learning library},
        author = {Paszke, Adam and Gross, Sam and Massa, Francisco and Lerer, Adam and Bradbury, James and Chanan, Gregory and Killeen, Trevor and Lin, Zeming and Gimelshein, Natalia and Antiga, Luca and others},
        booktitle = nips,
	volume = {32},
        pages = {8026-8037},
        month = Dec,
        year = {2019},
        publisher = publisher-curran,
        address = city-vancouver
}

@article{Wang2023Adaptive,
	title = {Adaptive Shells for Efficient Neural Radiance Field Rendering},
	author = {Wang, Zian and Shen, Tianchang and Nimier-David, Merlin and Sharp, Nicholas and Gao, Jun and Keller, Alexander and Fidler, Sanja and M{\"u}ller, Thomas and Gojcic, Zan},
	journal = acm-tog,
	volume = {42},
	number = {6},
	pages = {1-15},
	month = Dec,
	year = {2023},
	publisher = publisher-acmlong,
	doi = {10.1145/3618390},
	url = {https://doi.org/10.1145/3618390}
}

@inproceedings{Wang2023Neural,
	title = {Neural Fields Meet Explicit Geometric Representations for Inverse Rendering of Urban Scenes},
	author = {Wang, Zian and Shen, Tianchang and Gao, Jun and Huang, Shengyu and Munkberg, Jacob and Hasselgren, Jon and Gojcic, Zan and Chen, Wenzheng and Fidler, Sanja},
	booktitle = ieee-c-icvprx,
	shortjournalproceedings = {2023 IEEE/CVF Conf. Comput. Vis. Pattern Recognit. (CVPR)},
	pages = {8370-8380},
	month = Jun,
	year = {2023},
	publisher = publisher-ieee-short,
	address = city-vancouver,
	doi = {10.1109/cvpr52729.2023.00809},
	url = {https://doi.org/10.1109/CVPR52729.2023.00809}
}

@inproceedings{Cao2023HexPlane,
	title = {{HexPlane}: A Fast Representation for Dynamic Scenes},
	author = {Cao, Ang and Johnson, Justin},
	booktitle = ieee-c-icvprx,
	shortjournalproceedings = {2023 IEEE/CVF Conf. Comput. Vis. Pattern Recognit. (CVPR)},
	pages = {130-141},
	month = Jun,
	year = {2023},
	publisher = publisher-ieee-short,
	address = city-vancouver,
	doi = {10.1109/cvpr52729.2023.00021},
	url = {https://doi.org/10.1109/CVPR52729.2023.00021}
}

@article{Chen20243DReconstruction,
	title = {{3D} Reconstruction with Fast Dipole Sums},
	author = {Chen, Hanyu and Miller, Bailey and Gkioulekas, Ioannis},
	journal = acm-tog,
	volume = {43},
	number = {6},
	pages = {1-19},
	month = Nov,
	year = {2024},
	publisher = publisher-acmlong,
	doi = {10.1145/3687914},
	url = {https://doi.org/10.1145/3687914}
}

@inproceedings{Xu2022PointNeRF,
	title = {Point-{NeRF}: Point-based Neural Radiance Fields},
	author = {Xu, Qiangeng and Xu, Zexiang and Philip, Julien and Bi, Sai and Shu, Zhixin and Sunkavalli, Kalyan and Neumann, Ulrich},
	booktitle = ieee-c-icvprx,
	shortjournalproceedings = {2022 IEEE/CVF Conf. Comput. Vis. Pattern Recognit. (CVPR)},
	pages = {5428-5438},
	month = Jun,
	year = {2022},
	publisher = publisher-ieee-short,
	address = city-neworleans,
	doi = {10.1109/cvpr52688.2022.00536},
	url = {https://doi.org/10.1109/CVPR52688.2022.00536}
}

@inproceedings{Chen2023NeuRBF,
	title = {{NeuRBF}: A Neural Fields Representation with Adaptive Radial Basis Functions},
	author = {Chen, Zhang and Li, Zhong and Song, Liangchen and Chen, Lele and Yu, Jingyi and Yuan, Junsong and Xu, Yi},
	booktitle = ieee-c-iccvx,
	shortjournalproceedings = {2023 IEEE/CVF Int. Conf. Comput. Vis. (ICCV)},
	pages = {4159-4171},
	month = Oct,
	year = {2023},
	publisher = publisher-ieee-short,
	address = city-paris,
	doi = {10.1109/iccv51070.2023.00386},
	url = {https://doi.org/10.1109/ICCV51070.2023.00386}
}

@inproceedings{Yu2021PlenOctrees,
	title = {{PlenOctrees} for Real-time Rendering of Neural Radiance Fields},
	author = {Yu, Alex and Li, Ruilong and Tancik, Matthew and Li, Hao and Ng, Ren and Kanazawa, Angjoo},
	booktitle = ieee-c-iccvx,
	shortjournalproceedings = {2021 IEEE/CVF Int. Conf. Comput. Vis. (ICCV)},
	pages = {5732-5741},
	month = Oct,
	year = {2021},
	publisher = publisher-ieee-short,
	address = city-montreal,
	doi = {10.1109/iccv48922.2021.00570},
	url = {https://doi.org/10.1109/ICCV48922.2021.00570}
}

@inproceedings{Takikawa2021Neural,
	title = {Neural Geometric Level of Detail: Real-time Rendering with Implicit {3D} Shapes},
	author = {Takikawa, Towaki and Litalien, Joey and Yin, Kangxue and Kreis, Karsten and Loop, Charles and Nowrouzezahrai, Derek and Jacobson, Alec and McGuire, Morgan and Fidler, Sanja},
	booktitle = ieee-c-icvprx,
	shortjournalproceedings = {2021 IEEE/CVF Conf. Comput. Vis. Pattern Recognit. (CVPR)},
	pages = {11353-11362},
	month = Jun,
	year = {2021},
	publisher = publisher-ieee-short,
	address = city-nashville,
	doi = {10.1109/cvpr46437.2021.01120},
	url = {https://doi.org/10.1109/CVPR46437.2021.01120}
}

@article{Martel2021Acorn,
	title = {Acorn: : adaptive coordinate networks for neural scene representation},
	author = {Martel, Julien N. P. and Lindell, David B. and Lin, Connor Z. and Chan, Eric R. and Monteiro, Marco and Wetzstein, Gordon},
	journal = acm-tog,
	volume = {40},
	number = {4},
	pages = {1-13},
	month = Jul,
	year = {2021},
	publisher = publisher-acmlong,
	doi = {10.1145/3450626.3459785},
	url = {https://doi.org/10.1145/3450626.3459785}
}

@inproceedings{Sharma2024Volumetric,
	title = {Volumetric Rendering with Baked Quadrature Fields},
	author = {Sharma, Gopal and Rebain, Daniel and Yi, Kwang Moo and Tagliasacchi, Andrea},
	booktitle = eccv,
	shortjournalproceedings = {Eur. Conf. Comput. Vis.},
	series = lncs,
	volume = {15121},
	pages = {275-292},
	month = Nov,
	year = {2024},
	publisher = publisher-springerswiss,
	doi = {10.1007/978-3-031-73036-8_16},
	url = {https://doi.org/10.1007/978-3-031-73036-8_16}
}

@article{Sharp2022Spelunking,
	title = {Spelunking the deep},
	author = {Sharp, Nicholas and Jacobson, Alec},
	journal = acm-tog,
	volume = {41},
	number = {4},
	pages = {1-16},
	month = Jul,
	year = {2022},
	publisher = publisher-acmlong,
	doi = {10.1145/3528223.3530155},
	url = {https://doi.org/10.1145/3528223.3530155}
}

@inproceedings{Li2023NerfAcc,
	title = {{NerfAcc}: Efficient Sampling Accelerates {NeRFs}},
	author = {Li, Ruilong and Gao, Hang and Tancik, Matthew and Kanazawa, Angjoo},
	booktitle = ieee-c-iccvx,
	shortjournalproceedings = {2023 IEEE/CVF Int. Conf. Comput. Vis. (ICCV)},
	pages = {18491-18500},
	month = Oct,
	year = {2023},
	publisher = publisher-ieee-short,
	address = city-paris,
	doi = {10.1109/iccv51070.2023.01699},
	url = {https://doi.org/10.1109/ICCV51070.2023.01699}
}

@inproceedings{Sun2022Direct,
	title = {Direct Voxel Grid Optimization: Super-fast Convergence for Radiance Fields Reconstruction},
	author = {Sun, Cheng and Sun, Min and Chen, Hwann-Tzong},
	booktitle = ieee-c-icvprx,
	shortjournalproceedings = {2022 IEEE/CVF Conf. Comput. Vis. Pattern Recognit. (CVPR)},
	pages = {5449-5459},
	month = Jun,
	year = {2022},
	publisher = publisher-ieee,
	address = city-neworleans,
	doi = {10.1109/cvpr52688.2022.00538},
	url = {https://doi.org/10.1109/CVPR52688.2022.00538}
}

@inproceedings{FridovichKeil2022Plenoxels,
	title = {Plenoxels: Radiance Fields without Neural Networks},
	author = {Fridovich-Keil, Sara and Yu, Alex and Tancik, Matthew and Chen, Qinhong and Recht, Benjamin and Kanazawa, Angjoo},
	booktitle = ieee-c-icvprx,
	shortjournalproceedings = {2022 IEEE/CVF Conf. Comput. Vis. Pattern Recognit. (CVPR)},
	pages = {5491-5500},
	month = Jun,
	year = {2022},
	publisher = publisher-ieee,
	address = city-neworleans,
	doi = {10.1109/cvpr52688.2022.00542},
	url = {https://doi.org/10.1109/CVPR52688.2022.00542}
}

@article{Mildenhall2021NeRF,
	title = {{NeRF}: Representing scenes as neural radiance fields for view synthesis},
	author = {Mildenhall, Ben and Srinivasan, Pratul P. and Tancik, Matthew and Barron, Jonathan T. and Ramamoorthi, Ravi and Ng, Ren},
	journal = acm-cacm,
	volume = {65},
	number = {1},
	pages = {99-106},
	month = Dec,
	year = {2021},
	publisher = publisher-acmlong,
	doi = {10.1145/3503250},
	url = {https://doi.org/10.1145/3503250}
}

@inproceedings{Barron2022MipNeRF360,
	title = {Mip-{NeRF} 360: Unbounded Anti-Aliased Neural Radiance Fields},
	author = {Barron, Jonathan T. and Mildenhall, Ben and Verbin, Dor and Srinivasan, Pratul P. and Hedman, Peter},
	booktitle = ieee-c-icvprx,
	shortjournalproceedings = {2022 IEEE/CVF Conf. Comput. Vis. Pattern Recognit. (CVPR)},
	pages = {5460-5469},
	month = Jun,
	year = {2022},
	publisher = publisher-ieee,
	address = city-neworleans,
	doi = {10.1109/cvpr52688.2022.00539},
	url = {https://doi.org/10.1109/CVPR52688.2022.00539}
}

@inproceedings{Barron2021MipNeRF,
	title = {Mip-{NeRF}: A Multiscale Representation for Anti-Aliasing Neural Radiance Fields},
	author = {Barron, Jonathan T. and Mildenhall, Ben and Tancik, Matthew and Hedman, Peter and Martin-Brualla, Ricardo and Srinivasan, Pratul P.},
	booktitle = ieee-c-iccvx,
	shortjournalproceedings = {2021 IEEE/CVF Int. Conf. Comput. Vis. (ICCV)},
	pages = {5835-5844},
	month = Oct,
	year = {2021},
	publisher = publisher-ieee,
	address = city-montreal,
	doi = {10.1109/iccv48922.2021.00580},
	url = {https://doi.org/10.1109/ICCV48922.2021.00580}
}

@inproceedings{Chen2022TensoRF,
	title = {{TensoRF}: Tensorial Radiance Fields},
	author = {Chen, Anpei and Xu, Zexiang and Geiger, Andreas and Yu, Jingyi and Su, Hao},
	booktitle = eccv,
	shortjournalproceedings = {Eur. Conf. Comput. Vis.},
	series = lncs,
	volume = {13692},
	pages = {333-350},
	year = {2022},
	address = city-telaviv,
	publisher = publisher-springerswiss,
	doi = {10.1007/978-3-031-19824-3_20},
	url = {https://doi.org/10.1007/978-3-031-19824-3_20}
}

@inproceedings{Barron2023ZipNeRF,
	title = {Zip-{NeRF}: Anti-Aliased Grid-Based Neural Radiance Fields},
	author = {Barron, Jonathan T. and Mildenhall, Ben and Verbin, Dor and Srinivasan, Pratul P. and Hedman, Peter},
	booktitle = ieee-c-iccvx,
	shortjournalproceedings = {2023 IEEE/CVF Int. Conf. Comput. Vis. (ICCV)},
	pages = {19640-19648},
	month = Oct,
	year = {2023},
	publisher = publisher-ieee,
	address = city-paris,
	doi = {10.1109/iccv51070.2023.01804},
	url = {https://doi.org/10.1109/ICCV51070.2023.01804}
}

@article{Li2018Differentiable,
	title = {Differentiable {Monte} {Carlo} ray tracing through edge sampling},
	author = {Li, Tzu-Mao and Aittala, Miika and Durand, Fr{\'e}do and Lehtinen, Jaakko},
	journal = acm-tog,
	volume = {37},
	number = {6},
	pages = {1-11},
	month = Dec,
	year = {2018},
	publisher = publisher-acmlong,
	doi = {10.1145/3272127.3275109},
	url = {https://doi.org/10.1145/3272127.3275109}
}

@article{Nicolet2021Large,
	title = {Large steps in inverse rendering of geometry},
	author = {Nicolet, Baptiste and Jacobson, Alec and Jakob, Wenzel},
	journal = acm-tog,
	volume = {40},
	number = {6},
	pages = {1-13},
	month = Dec,
	year = {2021},
	publisher = publisher-acmlong,
	doi = {10.1145/3478513.3480501},
	url = {https://doi.org/10.1145/3478513.3480501}
}

@book{AkenineMoller2018RealTimeRendering,
	title = {Real-Time Rendering},
	author = {Akenine-M{\"o}ller, Tomas and Haines, Eric and Hoffman, Naty and Pesce, Angelo and Iwanicki, Micha\l{} and Hillaire, S\'ebastien},
	edition = {4},
	year = {2018},
	publisher = {AK Peters/CRC Press},
	address = {Boca Raton, Florida, USA},
}

@book{Hughes2014CGPP,
	title = {Computer Graphics: Principles and Practice},
	author = {Hughes, John and van Dam, Andries and McGuire, Morgan and Sklar, David and Foley, James D. and Feiner, Steven and Akeley, Kurt},
 	edition = {3},
	year = {2014},
	publisher = {Addison-Wesley},
	url = {http://vig.pearsoned.com/store/product/1,1207,store-12521\_isbn-0321399528,00.html},
}

@inproceedings{Sigg2006GPUBased,
	title = {{GPU}-Based Ray-Casting of Quadratic Surfaces},
	author = {Sigg, Christian and Weyrich, Tim and Botsch, Mario and Gross, Markus},
	booktitle = {Symposium on Point-Based Graphics},
	shortjournalproceedings = {Symp. Point-based Graph.},
	year = {2006},
	publisher = publisher-eurass,
	address = city-boston,
	doi = {10.2312/SPBG/SPBG06/059-065},
	url = {http://diglib.eg.org/handle/10.2312/SPBG.SPBG06.059-065}
}

@inproceedings{Li2022AnAdjustable,
	title = {An Adjustable Farthest Point Sampling Method for Approximately-sorted Point Cloud Data},
	author = {Li, Jingtao and Zhou, Jian and Xiong, Yan and Chen, Xing and Chakrabarti, Chaitali},
	booktitle = ieee-w-sips,
	shortjournalproceedings = {2022 IEEE Work. Signal Process. Syst. (sips)},
	pages = {1-6},
	month = Nov,
	year = {2022},
	publisher = publisher-ieee-short,
	address = {Rennes, France},
	doi = {10.1109/sips55645.2022.9919246},
	url = {https://doi.org/10.1109/SiPS55645.2022.9919246}
}

@inproceedings{Bavoil2007Multifragment,
	title = {Multi-fragment effects on the {GPU} using thek-buffer},
	author = {Bavoil, Louis and Callahan, Steven P. and Lefohn, Aaron and Comba, Jo{\~a}o L. D. and Silva, Cl{\'a}udio T.},
	booktitle = acm-s-igg,
	shortjournalproceedings = {Proc. 2007 Symp. Interact. 3D Graph. Games},
	pages = {97-104},
	month = Apr-May,
	year = {2007},
	address = {Seattle, Washington},
	publisher = {ACM},
	doi = {10.1145/1230100.1230117},
	url = {https://doi.org/10.1145/1230100.1230117}
}

@article{MoenneLoccoz20243DGaussian,
	title = {{3D} {Gaussian} Ray Tracing: Fast Tracing of Particle Scenes},
	author = {Moenne-Loccoz, Nicolas and Mirzaei, Ashkan and Perel, Or and de Lutio, Riccardo and Martinez Esturo, Janick and State, Gavriel and Fidler, Sanja and Sharp, Nicholas and Gojcic, Zan},
	journal = acm-tog,
	volume = {43},
	number = {6},
	pages = {1-19},
	month = Nov,
	year = {2024},
	publisher = publisher-acmlong,
	doi = {10.1145/3687934},
	url = {https://doi.org/10.1145/3687934}
}

@inproceedings{Guo2023VMesh,
	title = {{VMesh}: Hybrid Volume-Mesh Representation for Efficient View Synthesis},
	author = {Guo, Yuan-Chen and Cao, Yan-Pei and Wang, Chen and He, Yu and Shan, Ying and Zhang, Song-Hai},
	booktitle = acm-c-siggraph-aisa-papers,
	shortjournalproceedings = {SIGGRAPH Asia 2023 Conf. Pap.},
	pages = {1-11},
	month = Dec,
	year = {2023},
	publisher = {ACM},
	address = city-sydney,
	doi = {10.1145/3610548.3618161},
	url = {https://doi.org/10.1145/3610548.3618161}
}

@article{Ren2025OctreeGS,
	title = {Octree-{GS}: Towards Consistent Real-time Rendering with {LOD}-Structured {3D} {Gaussian}s},
	author = {Ren, Kerui and Jiang, Lihan and Lu, Tao and Yu, Mulin and Xu, Linning and Ni, Zhangkai and Dai, Bo},
	journal = ieee-tpami,
	pages = {1-15},
	year = {2025},
	publisher = publisher-ieee,
	doi = {10.1109/tpami.2025.3568201},
	url = {https://doi.org/10.1109/TPAMI.2025.3568201}
}

@inproceedings{Liu2019Soft,
	title = {Soft Rasterizer: A Differentiable Renderer for Image-Based {3D} Reasoning},
	author = {Liu, Shichen and Chen, Weikai and Li, Tianye and Li, Hao},
	booktitle = ieee-c-iccv,
	shortjournalproceedings = {2019 IEEE/CVF Int. Conf. Comput. Vis. (ICCV)},
	pages = {7707-7716},
	month = Oct,
	year = {2019},
	publisher = publisher-ieee,
	address = city-seoul,
	doi = {10.1109/iccv.2019.00780},
	url = {https://doi.org/10.1109/ICCV.2019.00780}
}

@inproceedings{Chao2025Textured,
	title = {Textured {Gaussian}s for Enhanced {3D} Scene Appearance Modeling},
	author = {Chao, Brian and Tseng, Hung-Yu and Porzi, Lorenzo and Gao, Chen and Li, Tuotuo and Li, Qinbo and Saraf, Ayush and Huang, Jia-Bin and Kopf, Johannes and Wetzstein, Gordon and Kim, Changil},
	booktitle = ieee-c-icvprx,
	shortjournalproceedings = {2025 IEEE/CVF Conf. Comput. Vis. Pattern Recognit. (CVPR)},
	pages = {8964-8974},
	month = Jun,
	year = {2025},
	publisher = publisher-ieee-short,
	address = city-nashville,
	doi = {10.1109/cvpr52734.2025.00838},
	url = {https://doi.org/10.1109/CVPR52734.2025.00838}
}

@inproceedings{Fang2024MiniSplatting,
	title = {Mini-Splatting: Representing Scenes with a Constrained Number of {Gaussian}s},
	author = {Fang, Guangchi and Wang, Bing},
	booktitle = eccv,
	shortjournalproceedings = {Eur. Conf. Comput. Vis.},
	series = lncs,
	volume = {15135},
	pages = {165-181},
	year = {2024},
	publisher = publisher-springerswiss,
	doi = {10.1007/978-3-031-72980-5_10},
	url = {https://doi.org/10.1007/978-3-031-72980-5_10}
}

@inproceedings{RotaBulo2024Revising,
	title = {Revising Densification in {Gaussian} Splatting},
	author = {Rota Bul{\`o}, Samuel and Porzi, Lorenzo and Kontschieder, Peter},
	booktitle = eccv,
	shortjournalproceedings = {Eur. Conf. Comput. Vis.},
	series = lncs,
	volume = {15121},
	pages = {347-362},
	month = Nov,
	year = {2024},
	publisher = publisher-springerswiss,
	doi = {10.1007/978-3-031-73036-8_20},
	url = {https://doi.org/10.1007/978-3-031-73036-8_20}
}

@inproceedings{Lu2024ScaffoldGS,
	title = {Scaffold-{GS}: Structured {3D} {Gaussian}s for View-Adaptive Rendering},
	author = {Lu, Tao and Yu, Mulin and Xu, Linning and Xiangli, Yuanbo and Wang, Limin and Lin, Dahua and Dai, Bo},
	booktitle = ieee-c-icvprx,
	shortjournalproceedings = {2024 IEEE/CVF Conf. Comput. Vis. Pattern Recognit. (CVPR)},
	pages = {20654-20664},
	month = Jun,
	year = {2024},
	publisher = publisher-ieee-short,
	address = city-seattle,
	doi = {10.1109/cvpr52733.2024.01952},
	url = {https://doi.org/10.1109/CVPR52733.2024.01952}
}

@inproceedings{Xu2024TextureGS,
	title = {Texture-{GS}: Disentangling the Geometry and Texture for {3D} {Gaussian} Splatting Editing},
	author = {Xu, Tian-Xing and Hu, Wenbo and Lai, Yu-Kun and Shan, Ying and Zhang, Song-Hai},
	booktitle = eccv,
	shortjournalproceedings = {Eur. Conf. Comput. Vis.},
	series = lncs,
	volume = {15083},
	pages = {37-53},
	month = Oct,
	year = {2024},
	publisher = publisher-springerswiss,
	doi = {10.1007/978-3-031-72698-9_3},
	url = {https://doi.org/10.1007/978-3-031-72698-9_3}
}

@inproceedings{Rong2025GStex,
	title = {{GStex}: Per-Primitive Texturing of {2D} {Gaussian} Splatting for Decoupled Appearance and Geometry Modeling},
	author = {Rong, Victor and Chen, Jingxiang and Bahmani, Sherwin and Kutulakos, Kiriakos N. and Lindell, David B.},
	booktitle = ieee-c-wacvx,
	shortjournalproceedings = {2025 IEEE/CVF Winter Conf. Appl. Comput. Vis. (WACV)},
	pages = {3508-3518},
	month = Feb,
	year = {2025},
	publisher = publisher-ieee-short,
	address = city-tucson,
	doi = {10.1109/wacv61041.2025.00346},
	url = {https://doi.org/10.1109/WACV61041.2025.00346}
}

@article{Reiser2024Binary,
	title = {Binary Opacity Grids: Capturing Fine Geometric Detail for Mesh-Based View Synthesis},
	author = {Reiser, Christian and Garbin, Stephan and Srinivasan, Pratul and Verbin, Dor and Szeliski, Richard and Mildenhall, Ben and Barron, Jonathan and Hedman, Peter and Geiger, Andreas},
	journal = acm-tog,
	volume = {43},
	number = {4},
	pages = {1-14},
	month = Jul,
	year = {2024},
	publisher = publisher-acmlong,
	doi = {10.1145/3658130},
	url = {https://doi.org/10.1145/3658130}
}

@article{Ye2025When,
	title = {When {Gaussian} Meets Surfel: Ultra-fast High-fidelity Radiance Field Rendering},
	author = {Ye, Keyang and Shao, Tianjia and Zhou, Kun},
	journal = acm-tog,
	volume = {44},
	number = {4},
	pages = {1-15},
	month = Jul,
	year = {2025},
	publisher = publisher-acmlong,
	doi = {10.1145/3730925},
	url = {https://doi.org/10.1145/3730925}
}

@inproceedings{Liu2025Deformable,
	title = {Deformable Beta Splatting},
	author = {Liu, Rong and Sun, Dylan and Chen, Meida and Wang, Yue and Feng, Andrew},
	booktitle = {Proceedings of the Special Interest Group on Computer Graphics and Interactive Techniques Conference Conference Paper},
	pages = {1-11},
	month = Jul,
	year = {2025},
	publisher = {ACM},
	address = {},
	doi = {10.1145/3721238.3730716},
	url = {https://doi.org/10.1145/3721238.3730716}
}

@inproceedings{Kheradmand20243DGaussian,
	title = {{3D} {Gaussian} Splatting as {Markov} Chain {Monte} {Carlo}},
	author = {Kheradmand, Shakiba and Rebain, Daniel and Sharma, Gopal and Sun, Weiwei and Tseng, Jeff and Isack, Hossam and Kar, Abhishek and Tagliasacchi, Andrea and Yi, Kwang Moo},
        booktitle = nips,
	volume = {37},
        pages = {80965-80986},
        month = Dec,
        year = {2024},
        publisher = publisher-curran,
        address = city-vancouver,
        url = {https://neurips.cc/virtual/2024/poster/94984}
}

@article{Muller2022Instant,
	title = {Instant neural graphics primitives with a multiresolution hash encoding},
	author = {M{\"u}ller, Thomas and Evans, Alex and Schied, Christoph and Keller, Alexander},
	journal = acm-tog,
	volume = {41},
	number = {4},
	pages = {1-15},
	month = Jul,
	year = {2022},
	publisher = publisher-acmlong,
	doi = {10.1145/3528223.3530127},
	url = {https://doi.org/10.1145/3528223.3530127}
}

@inproceedings{Yariv2023BakedSDF,
	title = {{BakedSDF}: Meshing Neural {SDFs} for Real-Time View Synthesis},
	author = {Yariv, Lior and Hedman, Peter and Reiser, Christian and Verbin, Dor and Srinivasan, Pratul P. and Szeliski, Richard and Barron, Jonathan T. and Mildenhall, Ben},
	booktitle = acm-c-siggraph-cp,
	pages = {1-9},
	month = Jul,
	year = {2023},
	publisher = {ACM},
	address = city-losangeles,
	doi = {10.1145/3588432.3591536},
	url = {https://doi.org/10.1145/3588432.3591536}
}

@inproceedings{Hamdi2024GES,
	title = {{GES}: Generalized Exponential Splatting for Efficient Radiance Field Rendering},
	author = {Hamdi, Abdullah and Melas-Kyriazi, Luke and Mai, Jinjie and Qian, Guocheng and Liu, Ruoshi and Vondrick, Carl and Ghanem, Bernard and Vedaldi, Andrea},
	booktitle = ieee-c-icvprx,
	shortjournalproceedings = {2024 IEEE/CVF Conf. Comput. Vis. Pattern Recognit. (CVPR)},
	pages = {19812-19822},
	month = Jun,
	year = {2024},
	publisher = publisher-ieee,
	address = city-seattle,
	doi = {10.1109/cvpr52733.2024.01873},
	url = {https://doi.org/10.1109/CVPR52733.2024.01873}
}

@inproceedings{Dai2024Highquality,
	title = {High-quality Surface Reconstruction using {Gaussian} Surfels},
	author = {Dai, Pinxuan and Xu, Jiamin and Xie, Wenxiang and Liu, Xinguo and Wang, Huamin and Xu, Weiwei},
	booktitle = acm-c-siggraph-papers,
	volume = {35},
	pages = {1-11},
	month = Jul,
	year = {2024},
	publisher = {ACM},
	address = city-denver,
	doi = {10.1145/3641519.3657441},
	url = {https://doi.org/10.1145/3641519.3657441}
}

@inproceedings{Huang20242DGaussian,
	title = {{2D} {Gaussian} Splatting for Geometrically Accurate Radiance Fields},
	author = {Huang, Binbin and Yu, Zehao and Chen, Anpei and Geiger, Andreas and Gao, Shenghua},
	booktitle = acm-c-siggraph-papers,
	volume = {35},
	pages = {1-11},
	month = Jul,
	year = {2024},
	publisher = {ACM},
	address = city-denver,
	doi = {10.1145/3641519.3657428},
	url = {https://doi.org/10.1145/3641519.3657428}
}

@article{Kerbl20233DGaussian,
	title = {{3D} {Gaussian} Splatting for Real-Time Radiance Field Rendering},
	author = {Kerbl, Bernhard and Kopanas, Georgios and Leimkuehler, Thomas and Drettakis, George},
	journal = acm-tog,
	volume = {42},
	number = {4},
	pages = {1-14},
	month = Jul,
	year = {2023},
	publisher = publisher-acmlong,
	doi = {10.1145/3592433},
	url = {https://doi.org/10.1145/3592433}
}

@inproceedings{Zhang2025Neural,
  title={Neural Shell Texture Splatting: More Details and Fewer Primitives},
  author={Zhang, Xin and Chen, Anpei and Xiong, Jincheng and Dai, Pinxuan and Shen, Yujun and Xu, Weiwei},
  booktitle = ieee-c-iccvx,
month = Oct,
  pages={25229-25238},
  year={2025},	
publisher = publisher-ieee-short,
address = city-honolulu
}

@inproceedings{Svitov2025Billboard,
  title={Billboard splatting ({BBSplat}): Learnable textured primitives for novel view synthesis},
  author={Svitov, David and Morerio, Pietro and Agapito, Lourdes and Del Bue, Alessio},
  booktitle = ieee-c-iccvx,
month = Oct,
  pages={25029-25039},
  year={2025},
publisher = publisher-ieee-short,
address = city-honolulu
}

@inproceedings{Zhang2018TheUnreasonable,
	title = {The Unreasonable Effectiveness of Deep Features as a Perceptual Metric},
	author = {Zhang, Richard and Isola, Phillip and Efros, Alexei A. and Shechtman, Eli and Wang, Oliver},
	booktitle = ieee-c-icvprx,
	shortjournalproceedings = {2018 IEEE/CVF Conf. Comput. Vis. Pattern Recognit.},
	pages = {586-595},
	month = Jun,
	year = {2018},
	publisher = publisher-ieee-short,
	address = city-saltlake,
	doi = {10.1109/cvpr.2018.00068},
	url = {https://doi.org/10.1109/CVPR.2018.00068}
}

@article{papantonakis2025content,
  title={Content-Aware Texturing for Gaussian Splatting},
  author={Papantonakis, Panagiotis and Kopanas, Georgios and Durand, Fredo and Drettakis, George},
  journal={arXiv preprint arXiv:2512.02621},
  year={2025}
}
}

\clearpage
\setcounter{page}{1}
\maketitlesupplementary

\section{Overview}
We include additional implementation details, dataset details, experiment details, and discussion below.

\section{Implementation Details}

\subsection{Instant-NGP}

\paragraph{Architecture.} The original Instant-NGP implementation assumes inputs $x \in [0, 1]^d$. Later works apply it to large-scale scenes by contracting space to this hypercube~\cite{Barron2022MipNeRF360,Barron2023ZipNeRF,Zhang2025Neural}. We instead make minor modifications to the hash-grid in order to accept unbounded inputs. 
First, we use a hash-grid, even for the coarse levels. Second, the original hash function used by M\"uller et al.~\cite{Muller2022Instant} has structured collisions when evaluated over both positive and negative inputs. We modify the hash to accept negative integers by applying the function 
\begin{equation*}
    \texttt{MapPositive}(x) = \begin{cases}2x-1\quad\quad\quad x>0,\\-2x\quad\quad\quad \text{otherwise},\end{cases}
\end{equation*}
to the hash function inputs.

\paragraph{Anti-aliasing.} As mentioned in the main text, we additionally perform anti-aliasing on the hash-grid using the down-weighting strategy with grid weight decay proposed in ZipNeRF~\cite{Barron2023ZipNeRF}. We describe how we derive our down-weighting factor, which differs slightly from that of Barron et al.~\cite{Barron2023ZipNeRF}.

Let $t^*$ be the depth along the ray and $f$ to be the focal length. We model the projected pixel footprint as a 2D isotropic normal distribution $\mathcal{N}$ in world-space, with mean $x^* := \origin + t^* \dir$ and standard deviation $\frac{t^*}{f}$, that is parallel to the image plane. The integrated texture which we seek to approximate is $\mathbb{E}_{x \sim \mathcal{N}}\left[\mathcal{F}(x) \right].$

Estimating the MLP as a linear function, we have
\begin{align*}
    \mathbb{E}_{x \sim \mathcal{N}}\left[\mathcal{F}(x) \right] &\approx \mathcal{G}\left(\mathbb{E}_{x \sim \mathcal{N}}\left[\mathcal{H}(x)\right]\right).
\end{align*}
The hash-grid at level $\ell$, $\mathcal{H}_\ell$, is queried by hashing $s_\ell x$ and using trilinear interpolation. For a coarse but efficient estimate of $\mathbb{E}_{x\sim \mathcal{N}}\left[\mathcal{H}_\ell(s_\ell x) \right]$, we only query $\mathcal{H}_\ell$ at $s_\ell x^*$ and approximate the rest. More formally, we model
\begin{equation*}
\mathcal{H}_\ell(s_\ell x^* + \epsilon) \approx w(\epsilon)\mathcal{H}_\ell(s_\ell x^*) +  (1- w(\epsilon))\mathcal{H}_{\ell}(s_\ell x^* + \epsilon),
\end{equation*}
where $w(\epsilon)$ indicates a unit cube, i.e. $w(\epsilon) = 1$ if $||\epsilon||_{\infty} \le \frac{1}{2}$ and $w(\epsilon) = 0$ otherwise. We assume that the unknown values are zero in expectation. Hence,
\begin{align*}
    \mathbb{E}_{x\sim \mathcal{N}}\left[\mathcal{H}_\ell(s_\ell x) \right] &\approx \mathbb{E}_{x\sim \mathcal{N}}\left[w(s_\ell(x -x^*))\mathcal{H}_\ell(s_\ell x^*)\right]\\
    &\approx \erf\left(\frac{f}{2\sqrt{2}s_\ell t^*}\right)^2 \mathcal{H}_\ell(s_\ell x^*).
\end{align*}

Using the approximation $\erf(x)^2 \approx 1 - \exp\left(\frac{-4x^2}{\pi}\right)$ yields our down-weighting factor \begin{equation*}
    \Delta_{\ell} := 1 - \exp\left(-\frac{1}{2\pi} \left(\frac{f}{s_\ell t^*}\right)^2\right).
\end{equation*}

The above derivation makes several very crude approximations. It's likely that many down-weighting factors are adequate. More high-powered anti-aliasing techniques could be used, but we found that this simple down-weighting was already enough to remove noticeable artifacts.

Note that the MLP in the tinycudann library does not use bias terms. As we use ReLU activations, a natural consequence is that $\mathcal{G}(\vec{0}) = \vec{0}$. We choose to preserve this property as our derivation approximated $\mathcal{G}$ to be linear, but we observe that the neural texture is mostly grey in the background of large scenes, particularly the \textsc{Train} scene of the Tanks \& Temples dataset.

\paragraph{Other details.} Unlike prior work in which the sample positions are assumed to be non-differentiable~\cite{Muller2022Instant,Zhang2025Neural}, we compute the positional gradient from the bilinear interpolation into each hash-grid level.

Both the hash-grid and MLP are computed using 16-bit floating point numbers. The latter's quantization is particularly important in order to leverage tensor core operations on recent NVIDIA GPUs, which greatly accelerates the MLP pass. The gradients are accumulated within 32-bit floating point tensors to maintain accuracy during optimization. We use the Instant-NGP codebase for the MLP pass and implement our own hash-grid following the design specified above. When performing the backward pass, we use the same default loss scale of $128$ as Instant-NGP to reduce underflow. We report the nexels storage memory using two bytes for each Instant-NGP parameter and four bytes for all other parameters.

\subsection{Adaptive Density Control}

\paragraph{Initialization.} Given $N$ points $x_0, \ldots, x_{N-1}$, farthest point sampling iteratively samples the point which is farthest from the ones already sampled~\cite{Li2022AnAdjustable}. In other words, it returns an indexing $I_0, I_1, \ldots, I_{N-1}$, such that for any $0 < i \le N-1$,
\begin{align*}
    I_i &:= \argmax_{k \not \in\{I_0, \ldots, I_{i-1}\}} \min_{0 \le j < i} d\left(x_k, x_{I_j}\right).
\end{align*}
When initializing our model, we use farthest point sampling to sample the positions. For the scale, the nearest-neighbors initialization suggested by Kerbl et al.~\cite{Kerbl20233DGaussian} is not ideal for our representation as it leads to small primitives. Instead, we run nearest-neighbors across prefixes of the farthest point sampling order so that the initial scale for point $i$ with position $x_{I_i}$ is approximately $\min_{0 \le j < i} d\left(x_{I_i}, x_{I_j}\right)$. This implicitly leads to a hierarchy where points sampled early are larger while points sampled later are smaller. These smaller primitives can potentially be removed if the optimization finds that the larger primitives can replace them with texture alone.

\paragraph{Densification.} In the backwards pass of each iteration, we return the blended $L_1$ errors as defined in the work of Rota Bul\`o et al.~\cite{RotaBulo2024Revising}. These blended errors are a per-primitive heuristic quantity where a higher error suggests that the primitive should be split. We accumulate these errors across iterations by adding them, rather than performing primitive-wise maximums. Inspired by 3DGS-MCMC~\cite{Kheradmand20243DGaussian}, we normalize the blended errors and use them as probability weights to sample $5\%$ of the current primitives for splitting.

Our splitting operation aims to preserve the textured surface. Say we are splitting surfel $i$. We will assume $\sigma_{i,1} \ge \sigma_{i,2}$ for convenience. The surfel is then split into two primitives both with scales $(\frac{\sigma_{i,1}}{2}, \sigma_{i,2})$ and with means $\mu_i - \frac{\sigma_{i,1}}{2} \mathbf{v}_{i,1}$ and $\mu_i + \frac{\sigma_{i,1}}{2} \mathbf{v}_{i,1},$ respectively. All other properties of the new primitives, including the $\gamma$ values, are copied from the original one. For $\gamma_i \to \infty$, this splits the quad perfectly in half along its long axis.

\section{Dataset Details}

We capture four scenes, \textsc{Graffiti}, \textsc{Grocery}, \textsc{Table}, and \textsc{Tripod}. All images were captured using a Canon EOS Rebel T7 equipped with an EF-S 18-55mm f/3.5-5.6 IS II zoom lens set to a focal length of approximately 35 mm. 

To ensure photometric consistency, we configured camera settings by capturing preliminary reference shots from multiple angles at each scene. We then locked these settings for the capture of each scene, fixing ISO, white balance, zoom level, and exposure per scene. Auto-focus was enabled. 
These scenes were selected specifically to evaluate reconstruction performance on high-frequency textures. A similar dataset was captured by Chao et al.~\cite{Chao2025Textured}, but the data was not made public for licensing reasons.

For the \textit{indoor} scenes (\textsc{Grocery}, \textsc{Table}, \textsc{Tripod}), we captured  360-degree concentric trajectories centered on the object of interest.  For the \textit{outdoor} scene (\textsc{Graffiti}), which lacks a singular central subject, we densely captured the region of interest. It took approximately 30 minutes to capture up to 300 images for each scene. After capturing, we manually removed images with any noticeable motion blur. We include brief descriptions of each scene below. The dataset will be made public.

\begin{table*}[t!]
	\small
\resizebox{\linewidth}{!}{
  \setlength{\tabcolsep}{4pt}
\begin{tabular}{lccccccccccccccc}
    \toprule
	& \multicolumn{5}{c}{Mip-NeRF360 Indoor} & \multicolumn{5}{c}{Mip-NeRF360 Outdoor} & \multicolumn{5}{c}{Custom}\\
	& PSNR & SSIM & LPIPS & Memory & \#Prim.
	& PSNR & SSIM & LPIPS & Memory & \#Prim.
    & PSNR & SSIM & LPIPS & Memory & \#Prim.\\
    \cmidrule(l{3pt}r{3pt}){1-1}\cmidrule(l{3pt}r{3pt}){2-6}\cmidrule(l{3pt}r{3pt}){7-11}\cmidrule(l{3pt}r{3pt}){12-16}
        3DGS &  30.40 & 0.920 & 0.190 & 290MB & 1.25M & 24.64 & 0.731 & 0.234 & 900MB & 3.88M & 26.65 & 0.858 & 0.224 & 1630MB & 7.03M\\
        2DGS &  30.41 & 0.917 & 0.192 & 180MB & 0.78M & 24.34 & 0.725 & 0.246 & 710MB & 3.06M & 25.98 & 0.831 & 0.267 & 420MB & 1.80M\\
        MiniSplatting &  30.43 & 0.922 & 0.188 & 90MB & 0.39M & 24.72 & 0.741 & 0.242 & 130MB & 0.57M & 24.88 & 0.803 & 0.297 & 100MB & 0.42M\\
        TriSplat &  30.80 & 0.928 & 0.160 & 530MB & 2.25M & 24.24 & 0.722 & 0.217 & 1080MB & 4.56M & 24.28 & 0.780 & 0.275 & 640MB & 2.73M\\
        NeST-Splat &  30.35 & 0.909 & 0.176 & 280MB & 0.33M & 23.74 & 0.703 & 0.242 & 580MB & 1.64M & 25.39 & 0.824 & 0.246 & 900MB & 3.24M\\
    \cmidrule(l{3pt}r{3pt}){1-1}\cmidrule(l{3pt}r{3pt}){2-6}\cmidrule(l{3pt}r{3pt}){7-11}\cmidrule(l{3pt}r{3pt}){12-16}
        3DGS-MCMC (400K) &  30.13 & 0.907 & 0.225 & 90MB & 0.40M & 24.03 & 0.672 & 0.341 & 90MB & 0.40M & 25.61 & 0.807 & 0.327 & 93MB & 0.40M\\
        3DGS-MCMC (1M) &  31.16 & 0.926 & 0.189 & 230MB & 1.00M & 24.74 & 0.722 & 0.272 & 230MB & 1.00M & 25.96 & 0.823 & 0.295 & 232MB & 1.00M\\
        3DGS-MCMC &  31.34 & 0.929 & 0.181 & 340MB & 1.45M & 25.04 & 0.744 & 0.232 & 830MB & 3.57M & - & - & - & - & -\\  
    \cmidrule(l{3pt}r{3pt}){1-1}\cmidrule(l{3pt}r{3pt}){2-6}\cmidrule(l{3pt}r{3pt}){7-11}\cmidrule(l{3pt}r{3pt}){12-16}
        DBS (400K) &  31.25 & 0.924 & 0.192 & 40MB & 0.40M & 24.11 & 0.692 & 0.312 & 40MB & 0.40M & 26.39 & 0.828 & 0.305 & 40MB & 0.40M\\
        DBS (1M) &  31.98 & 0.933 & 0.174 & 110MB & 1.00M & 24.56 & 0.727 & 0.257 & 110MB & 1.00M & 26.72 & 0.844 & 0.273 & 110MB & 1.00M\\ 
        DBS &  32.18 & 0.935 & 0.169 & 160MB & 1.50M & 24.83 & 0.745 & 0.209 & 470MB & 4.40M & 26.22 & 0.842 & 0.259 & 270MB & 2.50M\\ 
    \cmidrule(l{3pt}r{3pt}){1-1}\cmidrule(l{3pt}r{3pt}){2-6}\cmidrule(l{3pt}r{3pt}){7-11}\cmidrule(l{3pt}r{3pt}){12-16}
        BBSplat (40K) &  28.85 & 0.886 & 0.228 & 30MB & 0.04M & 22.39 & 0.592 & 0.359 & 30MB & 0.04M & 24.98 & 0.794 & 0.321 & 40MB & 0.04M\\
        BBSplat (100K) &  30.34 & 0.918 & 0.184 & 80MB & 0.10M & 23.17 & 0.647 & 0.304 & 80MB & 0.10M & 25.84 & 0.829 & 0.266 & 80MB & 0.10M\\
        BBSplat (400K) &  31.16 & 0.926 & 0.170 & 260MB & 0.40M & 23.71 & 0.675 & 0.273 & 260MB & 0.40M & 26.57 & 0.844 & 0.225 & 260MB & 0.40M\\
    \cmidrule(l{3pt}r{3pt}){1-1}\cmidrule(l{3pt}r{3pt}){2-6}\cmidrule(l{3pt}r{3pt}){7-11}\cmidrule(l{3pt}r{3pt}){12-16}
        Nexels (40K, $T=2^{19}$) &  29.19 & 0.894 & 0.199 & 40MB & 0.04M & 23.26 & 0.641 & 0.294 & 40MB & 0.04M & 26.20 & 0.825 & 0.261 & 40MB & 0.04M\\
        Nexels (40K) &  29.38 & 0.900 & 0.190 & 80MB & 0.04M & 23.25 & 0.647 & 0.282 & 80MB & 0.04M & 26.50 & 0.836 & 0.240 & 80MB & 0.04M\\
        Nexels (100K) &  29.94 & 0.906 & 0.183 & 90MB & 0.10M & 23.81 & 0.675 & 0.258 & 90MB & 0.10M & 26.62 & 0.844 & 0.237 & 90MB & 0.10M\\
        Nexels (400K) &  30.94 & 0.921 & 0.164 & 160MB & 0.40M & 24.48 & 0.707 & 0.230 & 160MB & 0.40M & 27.04 & 0.851 & 0.222 & 160MB & 0.40M\\
        Nexels (1M) &  31.32 & 0.926 & 0.155 & 310MB & 1.00M & 24.77 & 0.722 & 0.215 & 310MB & 1.00M & - & - & - & - & - \\
    \bottomrule
\end{tabular}
	}
\caption{\textbf{Quantitative results.} We show full results of the data points from Figure 5 of the main paper, as well as additional experiment configurations.
}
\label{tab:table_all}
\end{table*}

\begin{table*}[t!]
	\small
\resizebox{\linewidth}{!}{
  \setlength{\tabcolsep}{4pt}
\begin{tabular}{lccccccccccccccc}
    \toprule
	& \multicolumn{4}{c}{Mip-NeRF360 Indoor} & \multicolumn{5}{c}{Mip-NeRF360 Outdoor} & \multicolumn{2}{c}{Tanks \& Temples} & \multicolumn{4}{c}{Custom}\\
    & \textsc{Bns} & \textsc{Cnt} & \textsc{Ktc} & \textsc{Rm} & \textsc{Bcc} & \textsc{Flw} & \textsc{Grd} & \textsc{Stm} & \textsc{Trh} & \textsc{Trn} & \textsc{Trc} & \textsc{Grf} & \textsc{Grc} & \textsc{Tbl} & \textsc{Trp}\\
    \cmidrule(l{3pt}r{3pt}){1-1}\cmidrule(l{3pt}r{3pt}){2-5}\cmidrule(l{3pt}r{3pt}){6-10}\cmidrule(l{3pt}r{3pt}){11-12}\cmidrule(l{3pt}r{3pt}){13-16}
Nexels (40K, $T=2^{19}$) & 29.67 & 27.44 & 29.41 & 30.23 & 23.36 & 19.85 & 25.48 & 25.08 & 22.54 & - & - & 24.86 & 26.93 & 27.88 & 25.14\\
Nexels (40K)& 29.98 & 27.61 & 29.62 & 30.32 & 23.44 & 19.79 & 25.59 & 25.11 & 22.33 & - & - & 25.38 & 27.16 & 27.95 & 25.50\\
Nexels (100K)& 30.81 & 28.13 & 30.32 & 30.51 & 24.05 & 20.38 & 26.23 & 25.88 & 22.53 & - & - & 25.50 & 27.51 & 27.73 & 25.74\\
Nexels (400K)& 31.92 & 28.95 & 31.22 & 31.66 & 24.72 & 21.19 & 26.96 & 26.56 & 22.99 & 21.50 & 25.61 & 25.98 & 28.07 & 27.98 & 26.12\\
Nexels (1M)& 32.47 & 29.26 & 31.59 & 31.97 & 25.04 & 21.48 & 27.36 & 27.01 & 22.97 & - & - & - & - & - & -\\
    \cmidrule(l{3pt}r{3pt}){1-1}\cmidrule(l{3pt}r{3pt}){2-5}\cmidrule(l{3pt}r{3pt}){6-10}\cmidrule(l{3pt}r{3pt}){11-12}\cmidrule(l{3pt}r{3pt}){13-16}
Nexels (40K, $T=2^{19}$) & 0.907 & 0.875 & 0.895 & 0.899 & 0.645 & 0.488 & 0.784 & 0.688 & 0.600 & - & - & 0.743 & 0.853 & 0.901 & 0.802\\
Nexels (40K)& 0.913 & 0.880 & 0.901 & 0.903 & 0.657 & 0.492 & 0.793 & 0.691 & 0.600 & - & - & 0.767 & 0.857 & 0.905 & 0.815\\
Nexels (100K)& 0.923 & 0.892 & 0.910 & 0.899 & 0.690 & 0.530 & 0.817 & 0.728 & 0.607 & - & - & 0.776 & 0.864 & 0.904 & 0.831\\
Nexels (400K)& 0.935 & 0.905 & 0.921 & 0.922 & 0.730 & 0.577 & 0.842 & 0.764 & 0.624 & 0.802 & 0.880 & 0.789 & 0.878 & 0.906 & 0.829\\
Nexels (1M)& 0.942 & 0.911 & 0.927 & 0.925 & 0.749 & 0.595 & 0.855 & 0.781 & 0.629 & - & - & - & - & - & -\\
    \cmidrule(l{3pt}r{3pt}){1-1}\cmidrule(l{3pt}r{3pt}){2-5}\cmidrule(l{3pt}r{3pt}){6-10}\cmidrule(l{3pt}r{3pt}){11-12}\cmidrule(l{3pt}r{3pt}){13-16}
Nexels (40K, $T=2^{19}$) & 0.211 & 0.214 & 0.154 & 0.218 & 0.304 & 0.376 & 0.178 & 0.290 & 0.322 & - & - & 0.338 & 0.264 & 0.164 & 0.277\\
Nexels (40K)& 0.199 & 0.205 & 0.146 & 0.210 & 0.288 & 0.364 & 0.164 & 0.282 & 0.309 & - & - & 0.302 & 0.248 & 0.151 & 0.261\\
Nexels (100K)& 0.188 & 0.192 & 0.136 & 0.215 & 0.261 & 0.334 & 0.145 & 0.249 & 0.302 & - & - & 0.318 & 0.234 & 0.146 & 0.251\\
Nexels (400K)& 0.170 & 0.176 & 0.123 & 0.186 & 0.227 & 0.298 & 0.124 & 0.216 & 0.286 & 0.197 & 0.112 & 0.274 & 0.222 & 0.146 & 0.245\\
Nexels (1M)& 0.161 & 0.167 & 0.114 & 0.179 & 0.208 & 0.281 & 0.110 & 0.197 & 0.277 & - & - & - & - & - & -\\
    \bottomrule
\end{tabular}
	}
\caption{\textbf{Per-scene results.} We provide per-scene PSNR, SSIM, and LPIPS scores for our nexels at various settings. Scene names are abbreviated to their first three consonants.}
\label{tab:ours_detailed}
\end{table*}

\begin{itemize}
    \item{\textsc{Graffiti} (254 images).} This outdoor scene depicts an alleyway where a concrete driveway separates a fence from a graffiti-covered building.

    \item{\textsc{Grocery} (278 images).} Captured indoors, this scene centers on a $0.5\times1$ meter dining table covered in kitchen rags. On top of the rags, the table features a variety of grocery items with text-heavy labels as well as an open book.

    \item{\textsc{Table} (165 images).} This scene was taken inside a single room, with the camera focusing on a small table around $1$ meter in length. Two open books, a newspaper, and a deck of cards are laid out on the table, along with miscellaneous other objects.

    \item{\textsc{Tripod} (210 images).} Arranged inside a small room, this scene consists of magazines, markers, and various boxes surrounding a large poster propped up on a tripod.
\end{itemize}

\section{Experiment Details}

\subsection{Baseline Details}
We evaluated the baselines at various settings in order to obtain data points at a varying amount of primitives and memory. For 3DGS~\cite{Kerbl20233DGaussian}, 2DGS~\cite{Huang20242DGaussian}, and TriSplat~\cite{Held2026Triangle}, there is no fine control over the number of primitives. We use their default densification settings with no notable failures. 
For methods that provide separate hyperparameter sets for \textit{outdoor} and \textit{indoor} scenes, we used the \textit{indoor} settings for our custom scenes. Additional details are provided below.

\paragraph{3DGS-MCMC~\cite{Kheradmand20243DGaussian}.}
The original paper uses the bicubic filter from the PIL library to downscale the images, specifically through the \texttt{resolution} flag. This produces smoother images than the ImageMagick downscaling which was used by Barron et al.~\cite{Barron2021MipNeRF} on the Mip-NeRF360 dataset. As a result, scripts yield slightly better photometric results when using the \texttt{resolution} flag for downscaling rather than using the \texttt{images} flag to specify the image directories downscaled by ImageMagick. For fairness, we re-evaluated 3DGS-MCMC using the same downscaling strategy as in the original 3D Gaussian Splatting work. When training for a low amount of splats, we initialize using farthest point sampling with the same number of points as our initialization~\cite{Li2022AnAdjustable}.

\paragraph{BBSplat~\cite{Svitov2025Billboard}.} 
For BBSplat, we found that their scripts are run at a different resolution than the majority of Gaussian splatting works. We re-ran the baseline using the same resolution as in 3D Gaussian Splatting. Additionally, we set a maximum primitive cap of 400K and limited the number of SfM points for initialization to 150K. For all scenes, we use their sky-box initialization that adds another 10K points in the background. For BBSplat, we were unable to reproduce the results for the \textsc{Stump} scene when run at the standard resolution, as it achieved a significantly lower PSNR. For this scene only, we instead ran the script at their resolution and rescaled the renders to our resolution with Lanczos resampling. We evaluate the storage memory of BBSplat after it performs dictionary-based compression on the textures.

\input{fig/artifacts}
\begin{table}
    \centering
    \small
    \begin{tabular}{lccc}
    \toprule
    $K$ & PSNR & SSIM & LPIPS \\
    \midrule
    0 & 24.81 & 0.663 & 0.415\\
    1 & 25.80 & 0.750 & 0.254\\
    2 & \textbf{25.98} & \textbf{0.759} & \textbf{0.241}\\
    3 & 25.76 & 0.756 & 0.247\\
    4 & 25.44 & 0.747 & 0.258\\
    \bottomrule
    \end{tabular}
    \caption{\textbf{Texture limit.} We train 40K primitives at different texture limits, $K$, on the Mip-NeRF360 scenes.}
    \label{tab:texture_limit}
\end{table}

\paragraph{DBS~\cite{Liu2025Deformable}.} Like 3DGS-MCMC, the results in the original DBS paper were obtained using downsampling with the \texttt{resolution} flag. We re-ran the method using Mip-NeRF360's downscaled images. Furthermore, the original DBS uses a scene-specific number of iterations that explicitly checks the test set as a stopping condition. For fairness, we use the same number of iterations for all scenes (i.e. 30K).
On our custom scenes, we set the primitive cap for DBS to 3 million primitives for \textsc{Graffiti}, \textsc{Grocery}, and \textsc{Tripod}, and to 1 million for \textsc{Table}.
We use 1 million primitives for \textsc{Table} because using 3 million resulted in unstable behavior, and the PSNR dropped significantly.

\paragraph{MiniSplatting~\cite{Fang2024MiniSplatting}.}We ran MiniSplatting with their recommended indoor and outdoor settings for each respective scene. To obtain the two low memory comparisons shown in Figure 7, we set the sampling factor to $0.2$ and $0.25$, respectively.

\paragraph{NeST-Splatting~\cite{Zhang2025Neural}.} 
The densification strategy in NeST-Splatting does not allow setting a maximum number of primitives. On our custom \textsc{Graffiti} scene, this resulted in an out-of-memory error even with a 48GB GPU. The results of NeST-Splatting on the custom dataset reported in the paper were obtained by doubling the interval between densification steps for the \textsc{Graffiti} scene alone. To attain the models shown in Figure 6 of the main paper, with exactly 40K primitives, we added a maximum number to their densification. We initialize from a COLMAP point cloud sampled using farthest point sampling~\cite{Li2022AnAdjustable} to 20K points, matching our initialization. The storage memory was computed using 16-bit floating point numbers for the hash-grid parameters. 

\subsection{Additional Results}
We list the data points from Figure 5 of the main paper in \Cref{tab:table_all}. We include per-scene metrics for our method at different settings in \Cref{tab:ours_detailed}.

We also conduct an experiment to determine the best value of $K$ in \Cref{tab:texture_limit}. The metrics sharply improve when adding in any texture at all (i.e. going from $K=0$ to $K=1$) and reach a peak at $K=2$. Surprisingly, increasing $K$ beyond 2 leads to degradation in quality. Empirically, we observe noise in the background of test renders at higher values of $K$, suggesting that primitives which were never textured at training views are textured at test views.

\input{fig/limitations}
\begin{table}
    \centering
    \begin{tabular}{ccc}
    \toprule
    Routine & Indoor & Outdoor\\
    \midrule
    Collection Pass & 11.8 ms & 8.1 ms\\
    Texturing Pass & 12.7 ms & 8.4 ms\\
    \midrule
    Total & 24.5 ms & 16.4 ms\\
    \bottomrule
    \end{tabular}
    \caption{\textbf{Detailed timings.} We show timings for each pass averaged across the Mip-NeRF360 indoor and outdoor scenes with 400K primitives.}    \label{tab:detailed_timings}
\end{table}

\section{Further Discussion}

\subsection{Texturing Strategy}
Limiting the number of textured primitives to $K$ has a number of consequences which we outline. Firstly, the rendering speed becomes a compromise between that of rendering with per-primitive colours and that of texturing every primitive. The runtime of the texturing step is determined by the number of pixels and the number of per-pixel primitives that are textured, which is fixed.

In the forwards pass, the textured surfels determine the majority of the appearance. However, the non-textured surfels still make small improvements to the render in cases where the texture's appearance is not sufficient. This is useful around areas with strong reflections and transparent materials, as our rendering formulation does not encapsulate full lighting effects.

In the backwards pass, allowing non-textured primitives to contribute to the render allows them to receive gradient flow during back-propagation. This is important for early training iterations where the gradient moves primitives towards the underlying scene geometry, regardless of whether the primitive was textured. The neural texture does not need to be rendered at multiple overlapping primitives to guide its optimization. Thanks to the volumetric nature of the neural field, its gradient flow is the same whether it is rendered by several coinciding surfels or by a single surfel with the same total weight.

\subsection{Metrics}
We primarily show LPIPS throughout the paper as we find that it best corresponds to the reconstruction quality of fine texture details, which is a weakness of Gaussian splatting. It also generally corresponds better to human preference, as that was its original purpose~\cite{Zhang2018TheUnreasonable}. We note that the LPIPS formula used by virtually all Gaussian splatting works has a minor discrepancy from the one used by Zhang et al.~\cite{Zhang2018TheUnreasonable}. Specifically, the official 3DGS repository scales its images to $[0, 1]$ before computing the LPIPS, while the LPIPS library used expects images scaled to $[-1, 1]$. Nonetheless, the scores returned using this appear to be reasonable for comparison purposes so long as all baselines are calculated in the same way. To be consistent with the rest of the field, we use the same LPIPS calculation as 3DGS.

\subsection{Limitations}

As mentioned in the main text, nexels are not able to run at interactive rates on mobile devices or low-end GPUs. We show timings of the individual rendering passes in \Cref{tab:detailed_timings}. The neural texture step currently uses tensor core operations and would be much slower if implemented in WebGL or OpenGL.

\Cref{fig:artifacts} includes qualitative examples of the limitations mentioned in the main paper, particularly unseen regions and motion blur. One more limitation is the geometric accuracy of the representation, particularly in background regions. If the geometry does not optimize or densify correctly, then the neural texture is of little help. We show two failure cases in \Cref{fig:limitations}. In the first, the sparse geometries are not able to optimize towards thin structures. In the second, the background geometries optimize to a local minimum, resulting in blurry textures.

\end{document}